\documentclass[twoside]{article}

\usepackage{amazon_paper}

\usepackage{booktabs}
\usepackage{amsfonts}
\usepackage{amsmath}
\usepackage{amssymb}
\usepackage{mathtools}
\usepackage{amsthm}
\usepackage{nicefrac}
\usepackage{subcaption}
\usepackage[capitalize,noabbrev]{cleveref}
\usepackage{wrapfig}
\usepackage{makecell}

\theoremstyle{plain}

\theoremstyle{definition}

\theoremstyle{remark}

\newcommand{\pp}[1]{\vspace{1.2mm}\noindent\textbf{#1.}}
\newcommand{\sparagraph}[1]{\vspace{0.0mm}\noindent\textbf{#1.}}

\newcommand{\ours}{\mbox{\textsc{ReSkill}}}

\newcommand{\tpm}[1]{{\fontsize{4.5}{5}\selectfont$\pm$#1}}

\definecolor{Gray}{gray}{0.92}

\definecolor{ourred}{HTML}{E65100}
\definecolor{ourblue}{HTML}{1565C0}

\definecolor{paleblue}{HTML}{7EA6E0}
\definecolor{palered}{HTML}{F19C99}
\definecolor{palegreen}{HTML}{A8D5A2}

\definecolor{plotred}{HTML}{f77189}
\definecolor{plotgreen}{HTML}{33b07a}
\definecolor{plotpurple}{HTML}{cc7af4}

\definecolor{stage1}{HTML}{34A853}
\definecolor{stage2}{HTML}{A680B8}
\definecolor{stage3}{HTML}{009999}

\definecolor{mainboxbg}{HTML}{FAFAFA}
\definecolor{mainboxborder}{HTML}{A1C6EA}
\definecolor{successgreen}{HTML}{2E7D32}
\definecolor{failred}{HTML}{C62828}

\definecolor{g-red}{RGB}{213, 66, 56}
\definecolor{g-green}{RGB}{49, 149, 79}

\usepackage{tikz}

\newcommand{\dbar}[2]{%
  \ifdim #1pt > 0pt
    \pgfmathsetmacro{\dbarratio}{min(1, #1/#2)}%
    \pgfmathsetmacro{\dbarw}{0.28*\dbarratio}%
    \pgfmathsetmacro{\dbarshade}{25 + 65*\dbarratio}%
    \makebox[0pt][l]{\,\tikz[baseline=-0.4ex]{\fill[black!\dbarshade] (0,0) rectangle (\dbarw cm, 0.10cm);}}%
  \fi
}

\usepackage{colortbl}
\usepackage{float}
\usepackage{multirow}
\usepackage{algorithm}
\usepackage{algorithmic}
\usepackage{enumitem}
\usepackage[most]{tcolorbox}
\usepackage{tabularx}
\newtcolorbox{templatebox}[1]{%
  enhanced,
  breakable,
  colback=white,
  colframe=black!65,
  colbacktitle=black!80,
  coltitle=white,
  boxrule=0.9pt,
  arc=2pt,
  left=6pt, right=6pt, top=6pt, bottom=6pt,
  title={#1},
  fonttitle=\bfseries,
  sharp corners,
  boxed title style={sharp corners, boxrule=0pt}
}

\newtcolorbox{casebox}[1][stage1color]{%
  colback=mainboxbg, colframe=#1, boxrule=0pt, leftrule=3pt,
  left=8pt, right=6pt, top=5pt, bottom=5pt,
  arc=0pt, outer arc=0pt, sharp corners
}
\definecolor{actionbg}{HTML}{EAECEF}
\newcommand{\action}[1]{\par\vspace{1pt}\hspace*{12pt}\colorbox{actionbg}{\strut\texttt{#1}}\vspace{1pt}}

\newcommand{\successbadge}{\tikz[baseline=-0.5ex]{\node[fill=successgreen!15, draw=successgreen, rounded corners=2pt, inner sep=2pt, line width=0.4pt]{\scriptsize\bfseries\textcolor{successgreen}{SUCCESS}};}}
\newcommand{\failbadge}{\tikz[baseline=-0.5ex]{\node[fill=failred!12, draw=failred, rounded corners=2pt, inner sep=2pt, line width=0.4pt]{\scriptsize\bfseries\textcolor{failred}{FAIL}};}}

\definecolor{stage1color}{HTML}{E65100}
\definecolor{stage2color}{HTML}{00695C}
\definecolor{stage3color}{HTML}{4A148C}
\definecolor{skillcolor}{HTML}{1565C0}

\title{\ours: Reconciling Skill Creation \\
with Policy Optimization in Agentic RL}

\author{%
\small{Zelin He$^{1,*,\dagger}$, Haotian Lin$^{2,*}$, Boran Han$^{3,\ddagger}$, Wei Zhu$^{3,\ddagger}$,\\[0.45em]
Haoyang Fang$^{3}$, Bernie Wang$^{3}$, Xuan Zhu$^{3}$, Runze Li$^{1}$, Matthew Reimherr$^{1,2}$}%
  \\[0.8em]
  {\fontsize{10pt}{11pt}\selectfont
  \href{https://github.com/amazon-science/reskill}{\raisebox{-0.15em}{\includegraphics[height=1em]{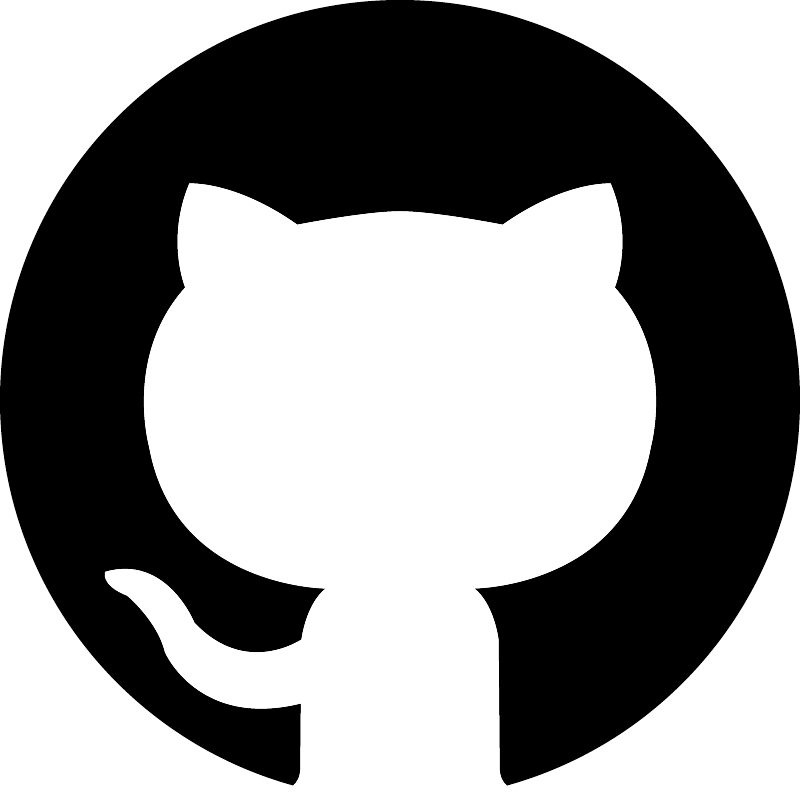}}\,Code}
  \qquad
  \href{https://amazon-science.github.io/reskill/}{\raisebox{-0.15em}{\includegraphics[height=1em]{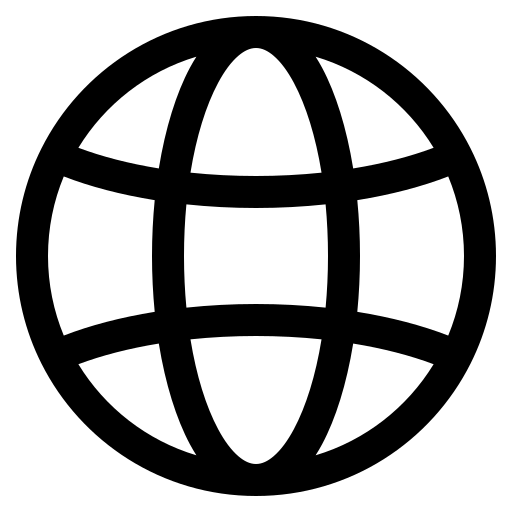}}\,Website}}
}

\begin{document}

\maketitle
\begingroup
\renewcommand{\thefootnote}{}
\footnotetext{\hspace{-1.8em}\begin{tabular}{@{}l@{}}
$^{*}$ Equal contribution. \quad $^{\ddagger}$ Project lead. \quad $^{\dagger}$ Work done during an internship at Amazon.\\[0.15em]
$^{1}$ The Pennsylvania State University \quad $^{2}$ Amazon IntelliHub \quad $^{3}$ AWS AI Labs
\end{tabular}}
\endgroup
\vspace{-1.0em}

\begin{abstract}
Agentic reinforcement learning (RL) enables LLM agents to improve continuously from environment rewards, yet the resulting policies do not systematically accumulate reusable strategies that generalize across tasks. Modular skills can provide such reusable strategies, yet existing skill-augmented RL methods decouple skill creation from policy optimization, risking adopting skills that conflict with the evolving policy. Inspired by Anthropic's Skill Creator, we introduce \ours, an RL-in-the-loop skill creation framework that reconciles skill evolution with policy learning. \ours~exploits the group-wise structure of GRPO to naturally embed three mechanisms with only marginal additional overhead: (1)~an assertion-driven skill creator that diagnoses failures from past experience and proposes conditional, trigger-based skill revisions; (2)~within-group rollout sampling that enables controlled comparison of skill versions, capturing which version best supports the policy's ongoing learning; and (3)~Thompson Sampling with adaptive discounting to balance exploration and exploitation in skill version selection as the policy evolves. Across several domains, \ours~consistently outperforms existing memory and skill-based RL methods, with the largest gains on unseen tasks. Analysis of the skill lifecycle shows skills being automatically created, tested, refined, and pruned as the policy improves, demonstrating reconciled skill-policy co-evolution.
\end{abstract}

\section{Introduction}

\begin{figure}[t]
\centering

\begin{minipage}[t]{0.67\textwidth}
    \centering
    \includegraphics[width=\linewidth]{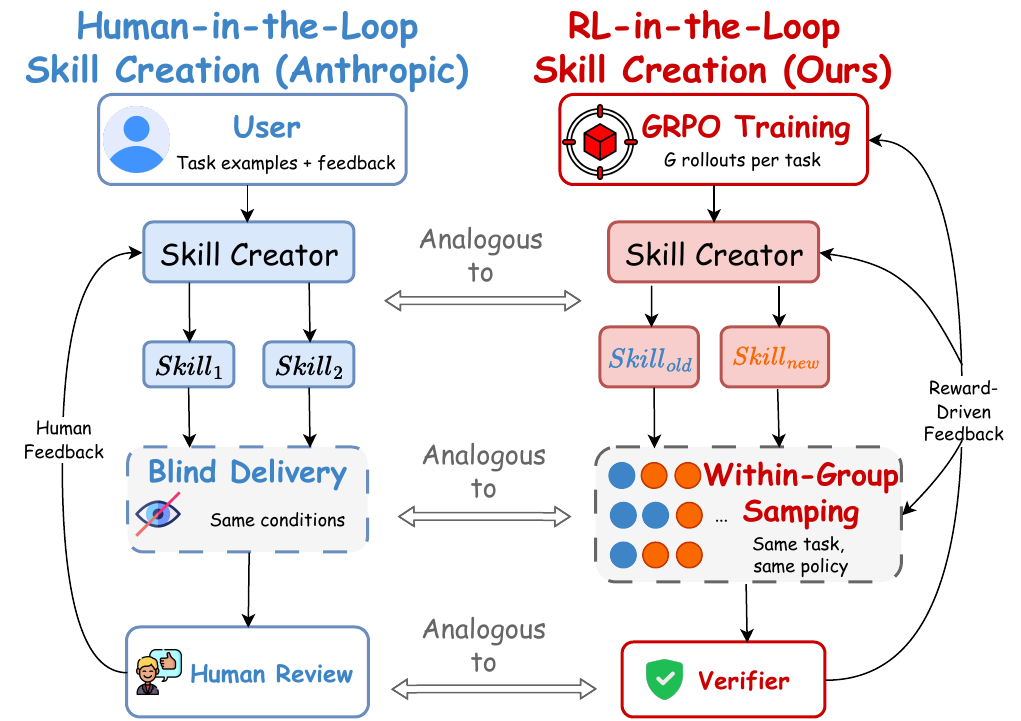}
    \vspace{2pt}
    {\small (a)}
\end{minipage}
\hfill
\begin{minipage}[t]{0.31\textwidth}
    \centering
    \includegraphics[width=\linewidth]{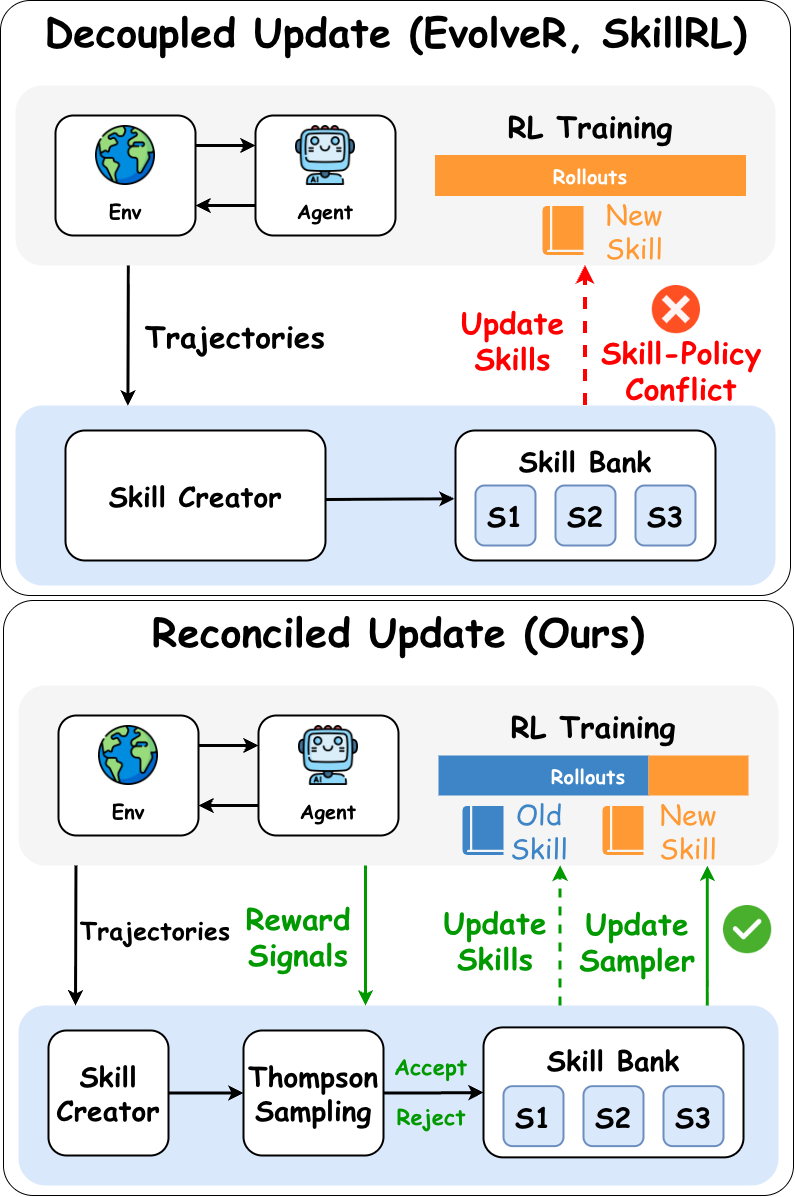}
    \vspace{2pt}
    {\small (b)}
\end{minipage}

\caption{
(a)~Inspired by Anthropic's human-in-the-loop Skill Creator, \ours~recasts skill creation as an RL-in-the-loop process.
(b)~Compared with decoupled skill-update methods, \ours~evaluates different skill versions inside the policy optimization loop and uses rewards for guided updates, achieving reconciled skill-policy updates.
}
\label{fig:intro_comparison}
\vspace{-8pt}
\end{figure}

Large language models (LLMs) have enabled autonomous agents capable of multi-turn interaction with complex environments \citep{yao2023react, liu2024agentbench}. Agentic reinforcement learning (RL), where agents are trained from verifiable environment rewards, has emerged as a promising paradigm \citep{jin2025searchr1, wei2025webagentr1, da2025agentrlvr}, yet the resulting policies do not systematically accumulate transferable problem-solving strategies across tasks. Recent work has shown that modular \textit{skills}, reusable snippets encoding task-specific strategies, can fill this gap \citep{wang2023voyager, zheng2025skillweaver, ni2026trace2skill}. A pioneering effort is Anthropic's Skill Creator \citep{anthropic2025skillcreator}, which automates skill creation through an iterative human-in-the-loop testing process (Figure~\ref{fig:intro_comparison}(a)). However, the Skill Creator operates on a static policy at inference time and relies on human feedback. A natural question arises: \textit{can we build an RL-in-the-loop skill creation process that dynamically creates and evaluates skills alongside an evolving policy?}

Recent approaches accumulate experiential knowledge during RL training through memory evolution \citep{zhang2026memrl}, principle distillation \citep{wu2025evolver}, and skill distillation \citep{xia2025skillrl}. However, they adopt external knowledge without principled testing, treating skill creation and policy learning as separate processes (Figure~\ref{fig:intro_comparison}(b)). Yet this approach risks skill-policy conflict, as externally produced skills may not align with the evolving policy. Ideally, whether a skill benefits learning should be assessed by testing it during policy training itself.

In this paper, we propose \ours, a framework that reconciles skill creation with policy learning. \ours~embeds the Anthropic Skill Creator cycle into the RL training loop and makes use of the group-wise structure of GRPO \citep{shao2024deepseekmath}: each training step samples a group of rollouts from the same task under the same policy, allowing us to assign competing skill versions within the same group and directly compare their effects. To orchestrate this allocation, Thompson Sampling \citep{thompson1933likelihood} with adaptive discounting is introduced to balance exploration and exploitation in skill version control as the policy evolves. Each rollout thus serves a \textit{triple role}: (i)~policy optimization via GRPO gradients, (ii)~failure diagnosis and skill revision through an assertion-driven creator adapted from Anthropic's Skill Creator, and (iii)~bandit evaluation that accepts or rejects skill versions based on how they support the policy's ongoing learning. Our method introduces marginal training overhead and requires no additional rollout budget.

We compare \ours~against existing memory and skill-based agentic RL baselines on embodied reasoning (ALFWorld) and agentic search (seven benchmarks) across multiple model scales, with additional experiments on coding (InterCode-SQL), web shopping (WANDS), and science discovery (ScienceWorld). \ours~achieves substantially improved performance, with the largest gains on unseen and hard tasks. The skill lifecycle analysis shows skills being automatically created, tested, refined, and pruned as the policy improves. Furthermore, the co-evolved policy can adapt to new domains (ALFWorld to ScienceWorld) at test time through skill creation alone, indicating that reconciled training develops a general capacity to follow and benefit from new skills.


\section{Preliminaries}

\pp{Agentic Reinforcement Learning with Skills}
We consider an LLM-based agent with policy $\pi_\theta$ that interacts with an environment over multiple turns via a tool engine $\mathcal{T}$. Given a task $q \sim \mathcal{D}$ and a skill bank $\mathcal{S}$ that provides strategic guidance with conditional loading, the agent generates a trajectory $\tau = (q,\, s_0, a_0, o_0,\, \ldots,\, s_K, a_K)$, where $a_t \sim \pi_\theta(\cdot \mid \tau_{<t}, s_t;\, \mathcal{T})$ is the agent's action, $o_t = \mathcal{T}(a_t)$ is the environment observation, and $s_t \subseteq \mathcal{S}$ is the subset of skills whose triggers fire. The learning objective maximizes the expected reward $r(\tau)$: $\max_{\pi_\theta} \mathbb{E}_{q \sim \mathcal{D},\, \tau \sim \pi_\theta(\cdot|q, \mathcal{S};\mathcal{T})} [ r(\tau) ]$.

\pp{Group Relative Policy Optimization (\textsc{Grpo})}
\textsc{Grpo} \citep{shao2024deepseekmath} optimizes the policy using group-relative advantages. For each task $q \sim \mathcal{D}$, \textsc{Grpo} samples $G$ trajectories $\{\tau_i\}_{i=1}^G$ from the current policy $\pi_{\theta}(\cdot \mid q, \mathcal{S}; \mathcal{T})$ and optimizes $\mathcal{J}_{\text{GRPO}}(\theta) = \mathbb{E}_{q \sim \mathcal{D},\; \tau_i \sim \pi_{\theta_{\text{old}}}(\cdot \mid q,\, \mathcal{S};\, \mathcal{T})} \left[ \mathcal{L}_{\text{GRPO}}(\theta;\, \{\tau_i\}_{i=1}^G) \right]$ with a static skill bank $\mathcal{S}$, where:
\begin{equation}
\mathcal{L}_{\text{GRPO}}(\theta;\, \{\tau_i\}_{i=1}^G) = \frac{1}{G} \sum_{i=1}^G \frac{1}{|\tau_i|} \sum_{t=1}^{|\tau_i|} \min \left( r_{i,t} \hat{A}_i,\; \text{clip}(r_{i,t}, 1{-}\epsilon, 1{+}\epsilon) \hat{A}_i \right) - \beta \mathbb{D}_{\text{KL}},
\label{eq:lgrpo}
\end{equation}
$r_{i,t}$ is the token-level importance sampling ratio, $\hat{A}_i$ is the group-normalized advantage, $\epsilon$ is the clip ratio, and $\beta$ controls the KL divergence penalty $\mathbb{D}_{\text{KL}}$ (details in Appendix~\ref{app:grpo}). All $G$ rollouts in a group share the same task $q$ and policy $\pi_\theta$, differing only in their stochastic trajectories, providing a natural controlled setting for comparing skill versions (\S\ref{sec:design}).


\section{Reconciling Skill Creation with Policy Learning}
\label{sec:method}

\begin{figure*}[!t]
\centering
\includegraphics[width=\textwidth]{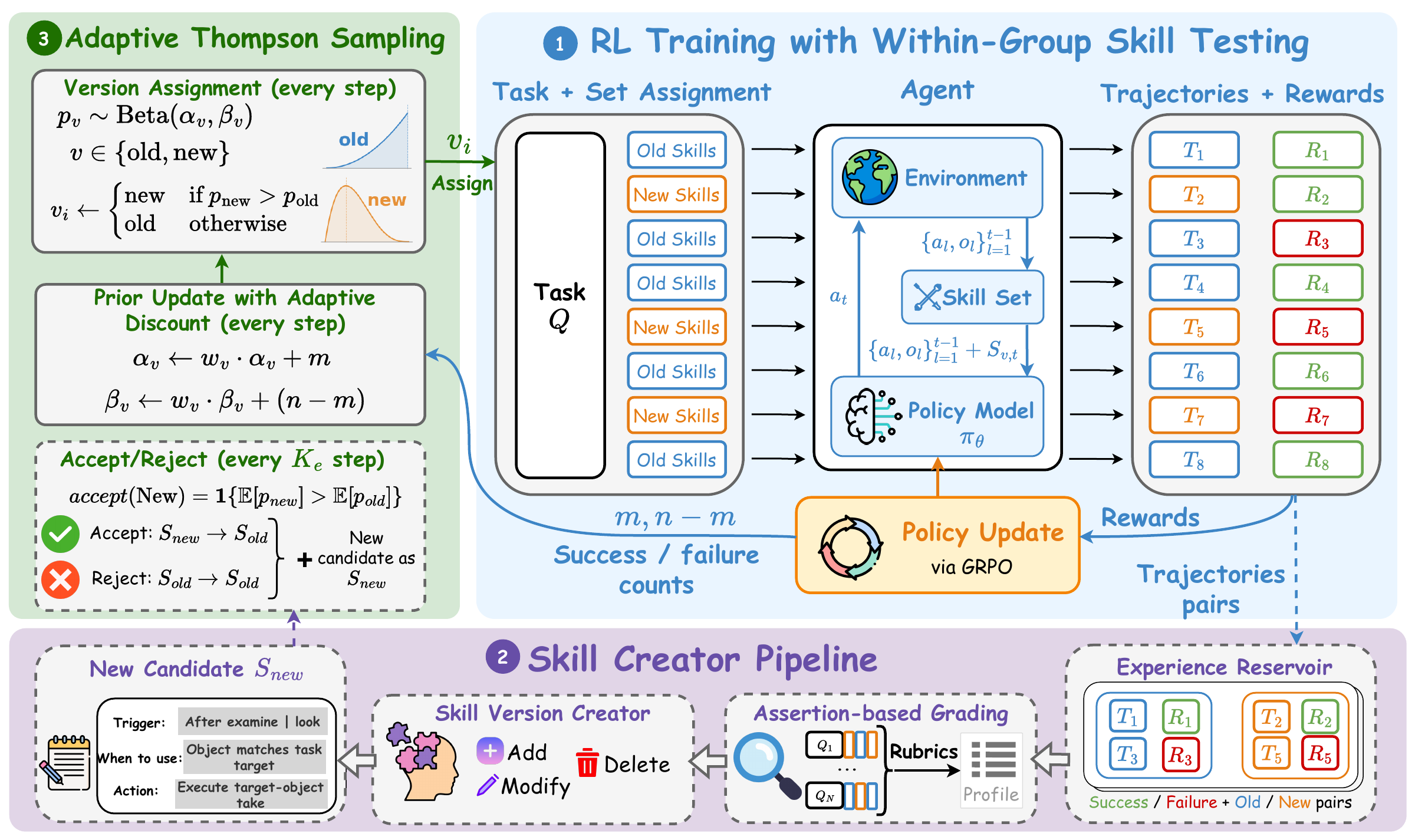}
\caption{Overview of \ours. \textbf{(1)}~RL training with within-group skill testing (\S\ref{sec:design}). \textbf{(2)}~RL-in-the-loop skill creation (\S\ref{sec:skill_creator}). \textbf{(3)}~RL-guided skill evolution with Thompson Sampling (\S\ref{sec:skill_evolution}).}
\label{fig:pipeline}
\vspace{-8pt}
\end{figure*}

Figure~\ref{fig:pipeline} illustrates the overall framework. We first describe how within-group skill testing integrates into \textsc{Grpo} training, enabling each rollout to simultaneously serve policy optimization, failure diagnosis, and skill evaluation (\S\ref{sec:design}). We then present the skill creator pipeline that produces candidate skill versions from ongoing training experience (\S\ref{sec:skill_creator}). Finally, we introduce Thompson Sampling to allocate rollouts between competing skill versions while maintaining training efficiency (\S\ref{sec:skill_evolution}). The complete procedure is summarized in Algorithm~\ref{alg:rerl}.

\subsection{RL Training with Within-Group Skill Testing}
\label{sec:design}

\pp{Within-Group Sampling}
To evaluate whether a skill benefits the evolving policy, we need to observe how the policy behaves under it. However, dedicated testing rollouts are expensive in multi-turn settings, and training under a single fixed skill bank loses adaptivity as skills change. We leverage GRPO's group structure to address both issues: for each rollout $i$ in a group, a version assignment $v_i$ is drawn from the Thompson Sampling posterior (\S\ref{sec:skill_evolution}), and the rollout is generated under the corresponding skill bank $\mathcal{S}_{v_i}$. The \ours~objective extends $\mathcal{J}_{\text{GRPO}}$ by replacing the static skill bank $\mathcal{S}$ with a version-dependent assignment:
\begin{equation}
\begin{aligned}
\mathcal{J}_{\text{\ours}}(\theta) = \mathbb{E}_{q \sim \mathcal{D},\; \textcolor{ourblue}{v_i \sim \mathrm{TS}(\alpha_v, \beta_v)},\; \tau_i \sim \pi_{\theta_{\text{old}}}(\cdot \mid q,\, \textcolor{ourblue}{\mathcal{S}_{v_i}};\, \mathcal{T})} \left[ \mathcal{L}_{\text{GRPO}}(\theta;\, \{\tau_i\}_{i=1}^G) \right],
\end{aligned}
\label{eq:rerl_objective}
\end{equation}
where $\mathcal{L}_{\text{GRPO}}$ is defined in Eq.~(\ref{eq:lgrpo}) and the \textcolor{ourblue}{blue} terms denote the components introduced by \ours: $\textcolor{ourblue}{v_i \sim \mathrm{TS}(\alpha_v, \beta_v)}$ assigns each rollout to either $\mathcal{S}_{\text{new}}$ or $\mathcal{S}_{\text{old}}$ via Thompson Sampling, and $\textcolor{ourblue}{\mathcal{S}_{v_i}}$ is the skill bank under the assigned version. Since all $G$ rollouts share the same task $q$ and policy $\pi_\theta$, the only variable is the skill version, enabling controlled comparison. At the same time, varying skills within the group builds the policy's capacity to reconcile with new skills.

\pp{Triple-Duty Rollouts}
This formulation allows each rollout to simultaneously serves three purposes without additional cost: (1)~its reward provides \textsc{Grpo} gradients for policy optimization; (2)~its completed trajectories feed an experience reservoir $\mathcal{R}$ that drives the skill creator (\S\ref{sec:skill_creator}); and (3)~its outcome under the assigned version updates the Thompson Sampling posteriors for the accept/reject decision (\S\ref{sec:skill_evolution}). The total rollout count remains identical to standard \textsc{Grpo}.

\subsection{RL-in-the-Loop Skill Creation}
\label{sec:skill_creator}

The within-group sampling (\S\ref{sec:design}) presents a cost free skill evaluation mechanism; this section describes the skill creator pipeline that produces candidate skill versions for evaluation. We adapt the diagnosis-proposal-validation loop inspired by Anthropic's Skill Creator \citep{anthropic2025skillcreator}, replacing human feedback with evolving policy trajectories and verifiable reward signals from the RL training loop. Full details are in Appendix~\ref{app:skill_pipeline} and~\ref{app:triggers}.

\pp{Experience Reservoir}
The original Skill Creator requires a human practitioner to supply execution examples for diagnosis. We replace this manual step with a continuously refreshed experience reservoir $\mathcal{R}$ that collects success-failure trajectory pairs from ongoing RL training, each trajectory is annotated with its skill version label $v_i$ and reward $r(\tau_i)$. Since $\mathcal{R}$ evolves alongside the policy, it always reflects the current behavioral landscape and provides the raw material for the diagnosis-proposal loop described below.

\pp{Conditional Skill Loading}
Each skill $S_k = (c_k, \text{trig}_k)$ is composed of structured content $c_k$ (situational applicability, action guidance, and counterexamples) with an iteratively refined trigger condition $\text{trig}_k$ parameterized by type. At each step, only skills whose triggers fire are loaded into the context, yielding the active subset of $\mathcal{S}$. Each loaded skill includes a \texttt{when\_to\_use} field that describes when its guidance should be followed, allowing the policy to decide whether the triggered skill is relevant to the current state.

\pp{Assertion-Based Failure Grading}
Rather than only diagnosing individual trajectories, we also build a general failure profile across the entire reservoir. A dynamic assertion set $\mathcal{A} = \{\phi_j\}$ of rule-based predicates is applied uniformly to all trajectories, yielding per-assertion pass rates $\hat{r}_j = |\mathcal{R}|^{-1}\sum_{\tau \in \mathcal{R}} \phi_j(\tau)$. An LLM-based analyzer receives these pass rates together with a stratified sample of trajectories and produces two outputs: (1)~a prevalence-ranked failure diagnosis that guides skill revision, and (2)~updates to $\mathcal{A}$ itself to keep the assertion set aligned with the policy's shifting failure landscape.

\pp{History-Informed Skill Revision}
Given the current skill bank, failure diagnosis with representative episodes, and the tracked history of prior proposals with their accept/reject outcomes, a skill creator agent proposes a candidate version $\mathcal{S}_{\text{new}}$: a coherent set of \textsc{add}, \textsc{modify}, and \textsc{delete} operations. Before entering testing, each proposed trigger's firing rate is validated against the reservoir to ensure sufficient activation coverage. We now have the newly proposed $\mathcal{S}_{\text{new}}$ and $\mathcal{S}_{\text{old}}$. In the next section, we introduce how rollouts are allocated between $\mathcal{S}_{\text{new}}$ and $\mathcal{S}_{\text{old}}$, and when to accept or reject.

\begin{wrapfigure}{r}{0.48\textwidth}
\vspace{-4pt}
\begin{minipage}{0.48\textwidth}
\refstepcounter{algorithm}
\vspace{4pt}
\hrule height 0.8pt
\vspace{3pt}
\small
\textbf{Algorithm \thealgorithm} \ \ours~(Full version in Appendix~\ref{app:method_details})
\label{alg:rerl_simple}
\vspace{3pt}
\hrule
\vspace{3pt}
\begin{algorithmic}[1]
    \STATE \textbf{Input}: Policy $\pi_{\theta_0}$, steps $T$, evolution freq.\ $K_e$.
    \FOR{$t=1, \dots, T$}
        \STATE \textcolor{paleblue}{[\textbf{Within-Group Skill Testing}]}
        \STATE Sample task $q$; draw skill version $v_i$. \hfill {\color{gray}\textit{// Eq.~(\ref{Thompson_Sampling})}}
        \STATE Generate trajectories $\{\tau_i\}_{i=1}^G$ under assigned skill $\mathcal{S}_{v_i}$; compute rewards $\{r_i\}$.
        \STATE Update $\pi_\theta$ by solving $\mathcal{J}_{\text{\ours}}(\theta)$. \hfill {\color{gray}\textit{// Eq.~(\ref{eq:rerl_objective})}}
        \STATE Add $\{(\tau_i, v_i, r_i)\}$ to experience reservoir $\mathcal{R}$.
        \STATE Update posterior $\text{Beta}(\alpha_v, \beta_v)$. \hfill {\color{gray}\textit{// Eq.~(\ref{eq:beta_update})}}
        \IF{$t \bmod K_e = 0$}
            \STATE \textcolor{palered}{[\textbf{Skill Evolution}]} Accept or reject $\mathcal{S}_{\text{new}}$.
            \STATE \textcolor{palegreen}{[\textbf{Skill Creation}]} Diagnose failures from $\mathcal{R}$; propose new $\mathcal{S}_{\text{new}}$.
            \STATE Reset posteriors; begin new test.
        \ENDIF
    \ENDFOR
\end{algorithmic}
\vspace{3pt}
\hrule height 0.8pt
\end{minipage}
\vspace{-8pt}
\end{wrapfigure}

\subsection{RL-Guided Skill Evolution with Thompson Sampling}
\label{sec:skill_evolution}

Unlike Anthropic's Skill Creator, which evaluates skills under a frozen policy, we evaluate how skills perform as the policy actively trains with them over multiple steps. This multi-step evaluation introduces two considerations: (1)~since the policy evolves during the evaluation window, earlier observations become increasingly stale and must be gradually discounted; and (2)~since every rollout also serves policy optimization, so allocation should favor the more promising version to maintain training efficiency. These considerations naturally motivate Thompson Sampling \citep{thompson1933likelihood} with adaptive discounting as the allocation mechanism.

\pp{Skill Version Assignment}
We treat the old and new skill versions as two arms, with success probabilities
$p_{\text{old}} \sim \operatorname{Beta}(\alpha_{\text{old}}, \beta_{\text{old}})$ and
$p_{\text{new}} \sim \operatorname{Beta}(\alpha_{\text{new}}, \beta_{\text{new}})$, both initialized as $\operatorname{Beta}(1,1)$ at the start of each evaluation cycle. Before generating the GRPO rollout group at training step~$t$, we estimate $\Pr(p_{\text{new}} > p_{\text{old}})$ by Monte Carlo sampling and define
\begin{equation*}
\pi_t(\text{new}) =
\operatorname{Clip}\!\left(
\Pr(p_{\text{new}} > p_{\text{old}}),\,
\varepsilon_{\text{ts}},\,
1-\varepsilon_{\text{ts}}
\right),
\qquad
\pi_t(\text{old}) = 1 - \pi_t(\text{new}).
\end{equation*}
Here, $\varepsilon_{\text{ts}}$ enforces a minimum level of exploration. We then assign versions independently within the GRPO group:
\begin{equation}
\label{Thompson_Sampling}
v_i \sim \operatorname{Categorical}\bigl(\{\text{new}: \pi_t(\text{new}),\; \text{old}: \pi_t(\text{old})\}\bigr),
\qquad i=1,\dots,G.
\end{equation}
As evidence accumulates across steps, the group-level allocation probability shifts toward the better-performing skill version while preserving controlled within-group comparison.

\pp{Policy-Aware Posterior Update}
As the policy evolves, recent steps provide the most informative signal about a skill version's current effectiveness, but a single step may lack sufficient samples for a reliable estimate. We therefore aggregate across the evaluation window but discount earlier evidence as the policy evolves. At each step~$t$, existing pseudo-counts are shrunk by $w_t$ before incorporating $m_t$ successes out of $n_t$ new episodes:
\begin{equation}
    \alpha_v \leftarrow w_t \cdot \alpha_v + m_t, \quad \beta_v \leftarrow w_t \cdot \beta_v + (n_t - m_t), \qquad w_t = (1 + n_t/M)^{-1},
    \label{eq:beta_update}
\end{equation}
where $v \in \{\text{new}, \text{old}\}$ and $M > 0$ is a memory parameter. Since Thompson Sampling allocates different sample sizes to each version, the discount adapts accordingly: a version with fewer observations ($n_t \ll M$) retains more prior evidence ($w_t \to 1$), while one with many observations ($n_t \gg M$) can aggressively discount the past ($w_t \to 0$).

The optimal value of $M$ varies across settings. We estimate $M$ from data via sequential predictive-likelihood maximization over completed evaluation cycles (Appendix~\ref{app:staleness_details}). At each cycle boundary, the system selects the $M$ that best explains the observed reward sequence under the discounted Beta model, requiring no manual tuning.

\pp{Accept/Reject Decision}
At the end of each cycle, the discounted posterior reflects how each version performed as the policy evolved with it. The candidate is accepted if $\mathbb{E}[p_{\text{new}}] > \mathbb{E}[p_{\text{old}}]$ under the discounted Beta posterior from Eq.~(\ref{eq:beta_update}). If accepted, $\mathcal{S}_{\text{new}}$ becomes the new baseline; otherwise the skill bank reverts to $\mathcal{S}_{\text{old}}$, and the next evolution cycle begins immediately.


\section{Experiments}
\label{sec:experiments}

\begin{table*}[!p]
\caption{Main results (mean\tpm{std} over 3 replicates, \%). $^\dagger$Per-cell best of BM25, HippoRAG, and long-context skill augmentation; $^\ddagger$per-cell best of ReMe and MemRL. Best per model scale \textbf{bolded}. Benchmark details in Appendix~\ref{app:benchmarks}.}
\label{tab:main_results}
\centering
\scriptsize
\renewcommand{\arraystretch}{0.92}
\setlength{\tabcolsep}{4pt}
\textbf{ALFWorld}\\[2pt]
\resizebox{\textwidth}{!}{%
\begin{tabular}{ll cccc @{\hskip 12pt} c @{\hskip 12pt} cccc @{\hskip 18pt} c}
\toprule
& & \multicolumn{4}{c}{Seen} && \multicolumn{4}{c}{Unseen} & \\
\cmidrule(lr){3-6} \cmidrule(lr){8-11}
& Method & Pick\&Place & Transform & Examine & Avg && Pick\&Place & Transform & Examine & Avg & Overall \\
\midrule
\textit{Proprietary} & Sonnet 4.5 & 89.8\tpm{0.0} & 73.5\tpm{2.1} & 96.2\tpm{5.4} & 82.5\tpm{1.5} && 73.2\tpm{4.2} & 72.0\tpm{0.0} & 100.0\tpm{0.0} & 76.1\tpm{1.3} & 79.2\tpm{1.5} \\
\midrule
\rowcolor{Gray} \multicolumn{12}{c}{\textbf{Qwen3-4B-Instruct}} \\
\midrule
\textit{Baseline} & ReAct & 54.7\tpm{0.0} & 28.1\tpm{0.0} & 28.6\tpm{0.0} & 37.1\tpm{0.0} && 30.8\tpm{0.0} & 28.9\tpm{0.0} & 66.7\tpm{0.0} & 32.8\tpm{0.0} & 35.0\tpm{0.0} \\
\midrule
\textit{w/ Evolve} & Best$^\ddagger$ & 60.5\tpm{1.6} & 44.6\tpm{0.7} & 79.5\tpm{3.6} & 54.5\tpm{0.9} && 45.5\tpm{3.0} & 58.2\tpm{2.3} & 55.6\tpm{7.9} & 52.7\tpm{0.4} & 53.6\tpm{0.5} \\
\textit{w/ Skill} & Best$^\dagger$ & 61.6\tpm{2.6} & 43.1\tpm{3.1} & 61.5\tpm{0.0} & 51.0\tpm{2.5} && 50.4\tpm{7.0} & 68.0\tpm{3.5} & 74.1\tpm{8.5} & 63.4\tpm{3.4} & 57.5\tpm{2.6} \\
\midrule
\multirow{6}{*}{\textit{w/ RL}} & GRPO & \textbf{93.8}\tpm{1.0} & 62.3\tpm{5.9} & 53.8\tpm{0.0} & 74.8\tpm{3.2} && 86.2\tpm{3.7} & 72.0\tpm{3.5} & 68.5\tpm{8.5} & 75.9\tpm{4.1} & 75.3\tpm{3.7} \\
& MemRL (+RL) & 89.8\tpm{2.9} & 75.8\tpm{1.9} & 71.8\tpm{4.4} & 81.2\tpm{0.4}\dbar{6.4}{15.2} && 78.9\tpm{2.8} & 76.6\tpm{0.5} & 94.4\tpm{5.6} & 79.1\tpm{1.5}\dbar{3.2}{15.2} & 80.2\tpm{0.6} \\
& EvolveR & 70.6\tpm{2.0} & 85.2\tpm{0.3} & 61.5\tpm{0.0} & 76.7\tpm{0.4}\dbar{1.9}{15.2} && 69.9\tpm{1.4} & 77.9\tpm{1.6} & 94.4\tpm{5.6} & 77.9\tpm{0.4}\dbar{2.0}{15.2} & 77.3\tpm{0.4} \\
& INSPO & 78.0\tpm{0.0} & 77.6\tpm{0.0} & 61.5\tpm{0.0} & 76.4\tpm{0.0}\dbar{1.6}{15.2} && 82.9\tpm{5.2} & 87.0\tpm{1.0} & 83.3\tpm{4.5} & 79.6\tpm{3.9}\dbar{3.7}{15.2} & 78.0\tpm{1.9} \\
& SkillRL & 89.8\tpm{0.0} & 85.9\tpm{0.0} & 61.5\tpm{0.0} & 85.7\tpm{0.0}\dbar{10.9}{15.2} && 69.9\tpm{1.4} & 86.9\tpm{1.3} & 88.9\tpm{0.0} & 82.1\tpm{1.3}\dbar{6.2}{15.2} & 83.9\tpm{0.6} \\
& \textbf{\ours} & 91.6\tpm{0.0} & \textbf{88.7}\tpm{0.0} & \textbf{76.9}\tpm{0.0} & \textbf{90.0}\tpm{0.0}\dbar{15.2}{15.2} && \textbf{87.4}\tpm{2.9} & \textbf{89.6}\tpm{1.9} & \textbf{96.3}\tpm{6.4} & \textbf{89.6}\tpm{0.8}\dbar{13.7}{15.2} & \textbf{89.8}\tpm{0.4} \\
\midrule
\rowcolor{Gray} \multicolumn{12}{c}{\textbf{Qwen3-8B}} \\
\midrule
\textit{Baseline} & ReAct & 73.4\tpm{0.0} & 44.8\tpm{0.0} & 64.3\tpm{0.0} & 59.3\tpm{0.0} && 69.2\tpm{0.0} & 65.0\tpm{0.0} & 66.7\tpm{0.0} & 64.9\tpm{0.0} & 62.0\tpm{0.0} \\
\midrule
\textit{w/ Evolve} & Best$^\ddagger$ & 79.1\tpm{4.0} & 58.3\tpm{2.5} & 61.5\tpm{0.0} & 66.2\tpm{2.8} && 57.7\tpm{7.0} & 76.4\tpm{1.7} & 57.4\tpm{6.9} & 65.7\tpm{1.2} & 65.9\tpm{1.8} \\
\textit{w/ Skill} & Best$^\dagger$ & 79.1\tpm{3.9} & 56.4\tpm{3.1} & 64.1\tpm{11.8} & 66.4\tpm{2.5} && 76.4\tpm{3.7} & 80.4\tpm{1.5} & 72.2\tpm{0.0} & 74.6\tpm{2.0} & 70.7\tpm{1.9} \\
\midrule
\multirow{6}{*}{\textit{w/ RL}} & GRPO & 93.8\tpm{1.0} & 72.1\tpm{5.3} & 66.7\tpm{4.4} & 80.7\tpm{1.9} && 82.1\tpm{1.4} & 81.8\tpm{3.4} & 79.6\tpm{3.2} & 81.6\tpm{2.3} & 81.1\tpm{1.7} \\
& MemRL (+RL) & 86.4\tpm{3.4} & 78.5\tpm{0.7} & 84.6\tpm{7.7} & 83.1\tpm{0.8}\dbar{2.4}{13.7} && 82.1\tpm{2.8} & 83.3\tpm{1.9} & 75.9\tpm{3.2} & 81.6\tpm{0.9}\dbar{0.0}{13.7} & 82.3\tpm{0.4} \\
& EvolveR & 88.7\tpm{2.6} & 79.8\tpm{2.1} & 74.4\tpm{4.4} & 83.3\tpm{0.4}\dbar{2.6}{13.7} && 70.7\tpm{2.4} & 89.2\tpm{0.2} & 83.3\tpm{5.6} & 82.8\tpm{1.3}\dbar{1.2}{13.7} & 83.1\tpm{0.9} \\
& INSPO & 90.4\tpm{1.6} & 79.2\tpm{1.6} & 71.8\tpm{3.6} & 83.2\tpm{1.6}\dbar{2.5}{13.7} && 88.1\tpm{2.0} & 89.2\tpm{1.5} & 81.5\tpm{6.9} & 87.8\tpm{1.4}\dbar{6.2}{13.7} & 85.5\tpm{1.1} \\
& SkillRL & 95.5\tpm{1.0} & \textbf{82.8}\tpm{3.2} & 87.2\tpm{4.4} & 89.0\tpm{0.4}\dbar{8.3}{13.7} && 82.9\tpm{2.4} & 78.3\tpm{2.1} & 94.4\tpm{5.6} & 82.6\tpm{1.7}\dbar{1.0}{13.7} & 85.8\tpm{0.8} \\
& \textbf{\ours} & \textbf{96.5}\tpm{1.0} & 81.4\tpm{2.2} & \textbf{94.9}\tpm{3.6} & \textbf{90.2}\tpm{0.7}\dbar{9.5}{13.7} && \textbf{99.3}\tpm{1.0} & \textbf{91.2}\tpm{2.4} & \textbf{100.0}\tpm{0.0} & \textbf{95.3}\tpm{1.3}\dbar{13.7}{13.7} & \textbf{92.7}\tpm{0.8} \\
\bottomrule
\end{tabular}}%
\vspace{4pt}

\textbf{Search}\\[2pt]
\resizebox{\textwidth}{!}{%
\begin{tabular}{ll ccc @{\hskip 12pt} c @{\hskip 12pt} cccccc @{\hskip 18pt} c}
\toprule
& & \multicolumn{3}{c}{Seen} && \multicolumn{6}{c}{Unseen} & \\
\cmidrule(lr){3-5} \cmidrule(lr){7-12}
& Method & NQ & HotpotQA & Avg && PopQA & TriviaQA & 2WikiMHQA & MuSiQue & Bamboogle & Avg & Overall \\
\midrule
\textit{Proprietary} & Sonnet 4.5 & 26.7\tpm{0.7} & 36.8\tpm{0.5} & 31.7\tpm{0.3} && 43.8\tpm{0.5} & 67.8\tpm{0.4} & 31.1\tpm{0.5} & 17.6\tpm{1.3} & 57.1\tpm{0.9} & 41.7\tpm{0.3} & 38.6\tpm{0.3} \\
\midrule
\rowcolor{Gray} \multicolumn{13}{c}{\textbf{Qwen3-4B-Instruct}} \\
\midrule
\textit{Baseline} & ReAct & 21.3\tpm{0.0} & 30.3\tpm{0.0} & 25.8\tpm{0.0} && 29.4\tpm{0.0} & 40.7\tpm{0.0} & 30.7\tpm{0.0} & 9.3\tpm{0.0} & 30.4\tpm{0.0} & 27.8\tpm{0.0} & 27.2\tpm{0.0} \\
\midrule
\textit{w/ Evolve} & Best$^\ddagger$ & 30.4\tpm{1.2} & 35.6\tpm{1.1} & 32.9\tpm{0.8} && 39.4\tpm{0.7} & 57.3\tpm{0.6} & 34.9\tpm{1.3} & 11.1\tpm{1.0} & 41.3\tpm{2.3} & 35.4\tpm{0.2} & 34.5\tpm{0.5} \\
\textit{w/ Skill} & Best$^\dagger$ & 31.6\tpm{0.5} & 36.0\tpm{0.3} & 33.6\tpm{0.1} && 42.1\tpm{1.5} & 55.1\tpm{1.0} & 30.9\tpm{2.1} & 10.8\tpm{1.0} & 42.1\tpm{0.5} & 34.9\tpm{1.5} & 34.5\tpm{1.0} \\
\midrule
\multirow{6}{*}{\textit{w/ RL}} & GRPO & 48.0\tpm{0.0} & 52.4\tpm{1.2} & 50.2\tpm{0.6} && 42.9\tpm{0.0} & 66.7\tpm{0.0} & 40.9\tpm{1.3} & \textbf{18.3}\tpm{0.7} & 40.3\tpm{2.8} & 42.0\tpm{0.5} & 44.6\tpm{0.3} \\
& MemRL (+RL) & 48.7\tpm{0.0} & 51.4\tpm{1.8} & 50.1\tpm{0.9} && 48.2\tpm{0.0} & 65.3\tpm{0.0} & 35.8\tpm{0.8} & 15.9\tpm{1.0} & 44.3\tpm{2.6} & 41.9\tpm{0.7} & 44.2\tpm{0.7} \\
& EvolveR & 49.3\tpm{0.0} & 49.8\tpm{0.4} & 49.6\tpm{0.2} && 46.6\tpm{0.4} & 65.7\tpm{0.0} & 43.4\tpm{1.1} & 16.7\tpm{1.5} & 43.2\tpm{2.1} & 43.1\tpm{0.3}\dbar{1.1}{3.4} & 45.0\tpm{0.2} \\
& INSPO & 48.6\tpm{0.2} & 53.6\tpm{0.6} & 51.1\tpm{0.3}\dbar{0.9}{3.4} && 46.5\tpm{0.0} & 63.0\tpm{0.0} & 44.0\tpm{2.0} & 15.8\tpm{0.2} & 42.4\tpm{1.7} & 42.3\tpm{0.5}\dbar{0.3}{3.4} & 45.1\tpm{0.3} \\
& SkillRL & 51.0\tpm{0.0} & 51.3\tpm{0.0} & 51.2\tpm{0.0}\dbar{1.0}{3.4} && 48.5\tpm{0.0} & 65.0\tpm{0.0} & 39.7\tpm{0.0} & 15.3\tpm{0.0} & 44.8\tpm{0.0} & 42.4\tpm{0.0}\dbar{0.4}{3.4} & 45.1\tpm{0.0} \\
& \textbf{\ours} & \textbf{51.6}\tpm{0.8} & \textbf{53.7}\tpm{1.2} & \textbf{52.6}\tpm{0.2}\dbar{2.4}{3.4} && \textbf{49.7}\tpm{0.2} & \textbf{66.9}\tpm{1.0} & \textbf{45.7}\tpm{0.6} & \textbf{18.3}\tpm{0.6} & \textbf{47.5}\tpm{0.9} & \textbf{45.4}\tpm{0.1}\dbar{3.4}{3.4} & \textbf{47.6}\tpm{0.1} \\
\midrule
\rowcolor{Gray} \multicolumn{13}{c}{\textbf{Qwen3-8B}} \\
\midrule
\textit{Baseline} & ReAct & 30.2\tpm{0.0} & 32.0\tpm{0.0} & 31.1\tpm{0.0} && 37.2\tpm{0.0} & 52.1\tpm{0.0} & 30.8\tpm{0.0} & 9.8\tpm{0.0} & 35.2\tpm{0.0} & 32.8\tpm{0.0} & 32.2\tpm{0.4} \\
\midrule
\textit{w/ Evolve} & Best$^\ddagger$ & 29.5\tpm{0.7} & 36.2\tpm{1.1} & 32.9\tpm{0.5} && 37.5\tpm{0.4} & 57.2\tpm{1.8} & 27.5\tpm{1.7} & 11.9\tpm{1.0} & 40.3\tpm{1.6} & 34.1\tpm{0.6} & 33.8\tpm{0.5} \\
\textit{w/ Skill} & Best$^\dagger$ & 32.8\tpm{0.5} & 39.2\tpm{1.3} & 36.0\tpm{0.4} && 42.6\tpm{0.0} & 57.0\tpm{0.9} & 30.6\tpm{1.3} & 11.2\tpm{0.2} & 40.8\tpm{1.6} & 35.5\tpm{0.6} & 35.2\tpm{0.5} \\
\midrule
\multirow{6}{*}{\textit{w/ RL}} & GRPO & 48.0\tpm{0.0} & 51.9\tpm{0.8} & 49.9\tpm{0.4} && 51.8\tpm{0.0} & 63.7\tpm{0.0} & 40.7\tpm{1.2} & 16.2\tpm{0.7} & 43.5\tpm{0.5} & 43.1\tpm{0.5} & 45.3\tpm{0.4} \\
& MemRL (+RL) & 48.3\tpm{0.0} & 54.8\tpm{0.3} & 51.6\tpm{0.2}\dbar{1.7}{4.9} && 49.8\tpm{0.0} & 65.7\tpm{0.0} & 41.2\tpm{0.9} & 19.0\tpm{0.5} & 42.1\tpm{2.6} & 43.8\tpm{0.4}\dbar{0.7}{4.9} & 46.2\tpm{0.3} \\
& EvolveR & \textbf{49.3}\tpm{0.0} & 53.1\tpm{0.3} & 51.2\tpm{0.2}\dbar{1.3}{4.9} && 51.2\tpm{0.0} & 65.7\tpm{0.0} & 43.9\tpm{1.3} & 16.7\tpm{0.0} & 45.6\tpm{0.0} & 44.5\tpm{0.3}\dbar{1.4}{4.9} & 46.6\tpm{0.1} \\
& INSPO & 48.7\tpm{0.0} & 55.2\tpm{0.8} & 51.9\tpm{0.4}\dbar{2.0}{4.9} && 48.2\tpm{0.0} & 64.7\tpm{0.0} & 39.4\tpm{1.5} & 16.9\tpm{0.4} & 42.1\tpm{2.3} & 42.3\tpm{0.2} & 45.0\tpm{0.1} \\
& SkillRL & 48.7\tpm{0.0} & 56.1\tpm{0.6} & 52.4\tpm{0.3}\dbar{2.5}{4.9} && 47.2\tpm{0.0} & 64.7\tpm{0.0} & 47.4\tpm{0.3} & 20.8\tpm{0.2} & 48.3\tpm{0.8} & 45.3\tpm{0.1}\dbar{2.2}{4.9} & 47.5\tpm{0.0} \\
& \textbf{\ours} & 49.0\tpm{0.0} & \textbf{58.3}\tpm{0.0} & \textbf{53.7}\tpm{0.0}\dbar{3.8}{4.9} && \textbf{52.3}\tpm{0.2} & \textbf{68.7}\tpm{0.6} & \textbf{47.6}\tpm{1.0} & \textbf{22.6}\tpm{1.3} & \textbf{50.4}\tpm{0.0} & \textbf{48.0}\tpm{0.2}\dbar{4.9}{4.9} & \textbf{49.8}\tpm{0.1} \\
\bottomrule
\end{tabular}}%
\vspace{4pt}

\includegraphics[width=\textwidth]{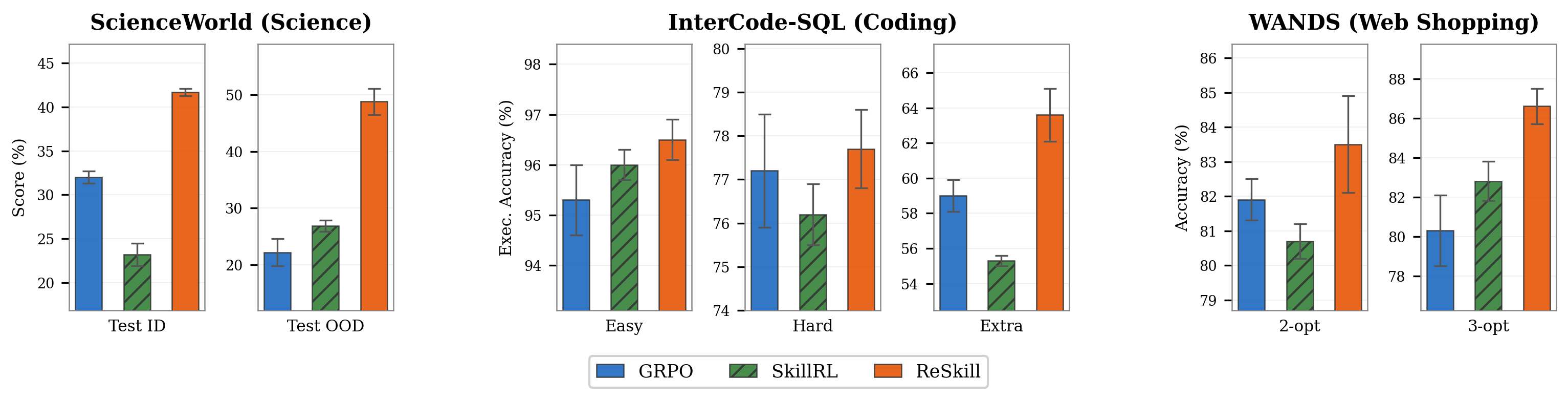}
\vspace{-6pt}
\captionof{figure}{Performance on additional benchmarks: \texttt{ScienceWorld} (electricity tasks), \texttt{InterCode-SQL}, and \texttt{WANDS}. The gap between \ours~and baselines widens consistently on harder or out-of-domain tasks.}
\label{fig:generalization_bars}
\end{table*}

\subsection{Setup}

\sparagraph{Environments}
We evaluate on five domains spanning embodied, retrieval, and tool-use reasoning.
\textit{(1)~Embodied reasoning}: \texttt{ALFWorld} \citep{ALFWorld20}, a text-based household environment evaluated on seen and unseen task configurations.
\textit{(2)~Agentic search}: agents answer factual questions via iterative search and reasoning. We train on \texttt{NQ} \citep{kwiatkowski2019natural} and \texttt{HotpotQA} \citep{yang2018hotpotqa}, and evaluate generalization on five unseen datasets: \texttt{PopQA} \citep{mallen2023popqa}, \texttt{TriviaQA} \citep{joshi2017triviaqa}, \texttt{2WikiMHQA} \citep{ho2020constructing}, \texttt{MuSiQue} \citep{trivedi2022musique}, and \texttt{Bamboogle} \citep{trivedi-etal-2023-interleaving}.
\textit{(3)~Science discovery, coding, and web shopping}: we further extend to \texttt{ScienceWorld} (electricity tasks) \citep{wang2022scienceworld}, \texttt{InterCode-SQL} \citep{yang2023intercode}, and \texttt{WANDS} \citep{chen2022wands} to assess broad applicability.

\pp{Baselines}
We compare \ours~against four categories of methods.
\textit{(1)~Baselines}: ReAct base model and closed-source model (Claude 4.5 Sonnet).
\textit{(2)~Evolvement methods}: memory and procedure evolution without policy training, including \textsc{Reme}~\citep{cao2025reme} and \textsc{Memrl}~\citep{zhang2026memrl}.
\textit{(3)~Skill-only baselines}: using the oracle skill bank from \textsc{Skillrl}~\citep{xia2025skillrl}, we test retrieval-based loading (\texttt{BM25}, \texttt{HippoRAG}~\citep{gutierrez2024hipporag}) and full skill loading (long-context) on the base model without RL.
\textit{(4)~RL-based methods}: \textsc{Grpo}~\citep{shao2024deepseekmath} (no skills), \textsc{Memrl}~(+RL), \textsc{Evolver}~\citep{wu2025evolver}, \textsc{Inspo}~\citep{zhou2025inspo}, and \textsc{Skillrl}.

\pp{Implementation Details}
We use Qwen3-4B-Instruct-2507 and Qwen3-8B~\citep{bai2023qwen} as base models; unless otherwise specified, results report the 4B model. For fair comparison, all evolve, skill, RL-based methods and \ours~use the same Claude 4.5 Sonnet for skill or memory creation (base-model alternative ablated in Table~\ref{tab:ablation}). For \ours, we set the evolution frequency $K_e{=}5$ and skill bank size to 8 (sensitivity analysis in Appendix~\ref{app:sensitivity}). Full training hyperparameters are in Appendix~\ref{app:implementation}.

\subsection{Main Results}

\begin{figure*}[!t]
\centering
\includegraphics[width=\textwidth]{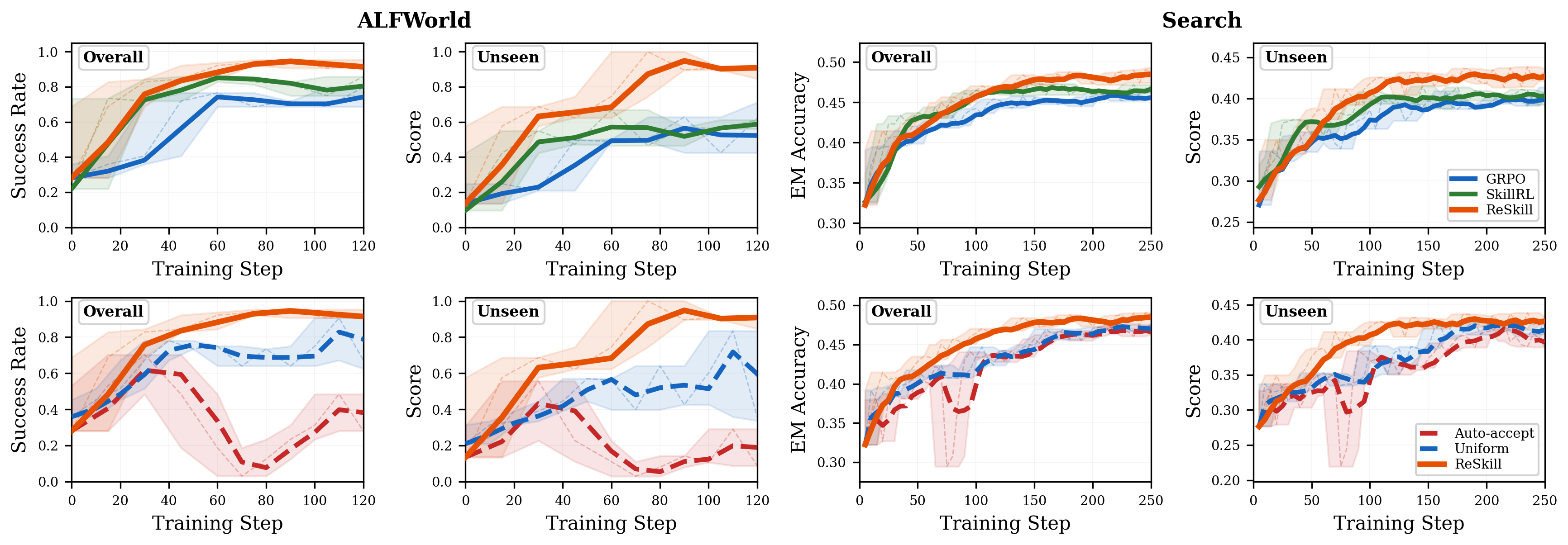}
\caption{Training dynamics on held-out validation subsets. \textbf{Top}: method comparison; \textbf{Bottom}: version evaluation strategy ablation (auto-accept: all proposed skills adopted without testing; uniform: equal allocation between versions without Thompson Sampling). Unseen splits report \emph{step-weighted score}~\citep{feng2025group}. Full test-set results in Tables~\ref{tab:main_results} and~\ref{tab:ablation}.}
\label{fig:training_curves}
\end{figure*}

Table~\ref{tab:main_results} summarizes performance across benchmarks, model scales, and seen/unseen splits; Figure~\ref{fig:training_curves} shows validation curves during training, comparing \ours~against \textsc{Grpo}, skill-augmented RL methods, and two key ablation variants.

\pp{Consistent Gains Across Benchmarks and Scales}
\ours~outperforms all baselines on both benchmarks at both model scales (Table~\ref{tab:main_results}). Non-RL methods trail by 22.0--32.3\% on \texttt{ALFWorld} and 13.1--14.6\% on \texttt{Search}, underscoring the importance of policy optimization; \ours~lifts over vanilla \textsc{Grpo} by 3.0--14.5\%, demonstrating the value of skill guidance. Most importantly, \ours~outperforms the strongest memory and skill-augmented RL baseline by 5.9\%/6.9\% on \texttt{ALFWorld} and 2.5\%/2.3\% on \texttt{Search} (4B/8B), a gap attributable to reconciling skill evolution with the evolving policy rather than treating the two as independent processes.

\pp{Amplified Gains on Unseen Tasks}
The gains are particularly strong on unseen tasks. On \texttt{ALFWorld} (8B), \ours~achieves 95.3\% on unseen tasks, a 12.7\% gap over \textsc{Skillrl} (82.6\%) that is substantially larger than its seen-split margin (1.2\%). On \texttt{Search}, \ours's improvement over \textsc{Skillrl} on unseen splits is about 2.1$\times$ larger than on seen splits. Figure~\ref{fig:training_curves} reveals that methods track closely in early training, but \ours~pulls ahead on unseen splits in later stages, presumably because reconciled evolution retains only skill versions that generalize better.

\pp{Broad Applicability}
The pattern of amplified gains on harder and out-of-domain tasks is further validated across three extension domains (Figure~\ref{fig:generalization_bars}). On \texttt{ScienceWorld}, \ours~reaches 48.8\% on OOD tasks, nearly doubling \textsc{Skillrl} (26.9\%), whose static skill bank degrades sharply on unseen task variants. On \texttt{InterCode-SQL}, methods perform comparably on Hard queries, but \ours~pulls ahead on Extra-hard (63.6\% vs.\ 55.3\% \textsc{Skillrl}), where compositional SQL reasoning benefits most from evolved skills. \texttt{WANDS} shows consistent gains across optimization horizons, with \ours~reaching 86.6\% on the harder 3-opt setting vs.\ \textsc{Skillrl}'s 82.1\%. Across all three domains, the gap between \ours~and the strongest baseline widens as task difficulty increases.

\subsection{Framework Analysis}
\label{sec:ablation}

\noindent\begin{minipage}[t]{0.47\textwidth}
\vspace{0pt}
\pp{Importance of Reconciled Evolution}
One key design in \ours~is evaluating whether a proposed skill version reconciles with the current policy before adopting it. Figure~\ref{fig:training_curves} (bottom row) makes this visible: the auto-accept curve (red), which injects every proposed skill without testing, rises initially but then collapses mid-training on \texttt{ALFWorld} and never recovers, with permanent damage to the final performance on both benchmarks (62.7\% and 45.3\%, Table~\ref{tab:ablation}). On \texttt{ALFWorld}, this falls \textit{below} vanilla \textsc{Grpo} (75.3\%), meaning untested skill injection actively harms the policy. The uniform allocation curve (blue) is more stable but remains behind \ours~(orange), as ineffective skill versions consume half the rollouts throughout the evaluation window,
\end{minipage}\hfill
\begin{minipage}[t]{0.49\textwidth}
\vspace{0pt}
\captionof{table}{Framework analysis (\%). Each variant modifies one component from the full \ours~system.}
\label{tab:ablation}
\centering
\footnotesize
\renewcommand{\arraystretch}{0.98}
\setlength{\tabcolsep}{3pt}
\begin{tabular*}{\linewidth}{@{\extracolsep{\fill}}l cc}
\toprule
\textbf{Variant} & \textbf{ALFWorld} & \textbf{Search} \\
\midrule
\textbf{\ours} & 89.8\tpm{0.4} & 47.6\tpm{0.1} \\
\midrule
\multicolumn{3}{l}{\textit{Co-Evolution Order}} \\
\quad Skill-first, then policy & 79.4\tpm{0.2} & 45.5\tpm{0.5} \\
\quad Policy-first, then skill & 71.3\tpm{1.8} & 45.9\tpm{0.0} \\
\midrule
\multicolumn{3}{l}{\textit{Skill Version Testing}} \\
\quad w/o testing (auto-accept) & 62.7\tpm{0.2} & 45.3\tpm{0.6} \\
\quad w/ uniform allocation & 73.2\tpm{1.7} & 46.0\tpm{0.0} \\
\midrule
\multicolumn{3}{l}{\textit{Evidence Aggregation ($w_t$, Eq.~\ref{eq:beta_update})}} \\
\quad Last-step only ($w_t{=}0$) & 76.9\tpm{1.0} & 46.7\tpm{0.2} \\
\quad Uniform ($w_t{=}1$) & 86.5\tpm{1.6} & 47.0\tpm{0.0} \\
\midrule
\multicolumn{3}{l}{\textit{Skill Creator}} \\
\quad w/o assertion-based analyzer & 84.7\tpm{1.3} & 46.7\tpm{0.1} \\
\quad w/ base-model skill creator & 85.0\tpm{0.5} & 47.1\tpm{0.1} \\
\bottomrule
\end{tabular*}
\end{minipage}

\vspace{2pt}
\noindent reducing sample efficiency and depressing the final result (46.0\% vs.\ 47.6\% on \texttt{Search}). \ours's Thompson Sampling design reconciles skill evolution with the policy by testing versions under real training conditions before adoption, avoiding both failure modes.

\pp{Robustness to Skill Creator Choice}
A natural concern is whether the improvements primarily reflect the stronger LLM (Claude 4.5 Sonnet) used for skill creation. To ablate this, we replace Claude with the base model itself (Qwen3-4B-Instruct) for skill creation. This incurs a 4.8\% drop on \texttt{ALFWorld} and only 0.5\% on \texttt{Search} (Table~\ref{tab:ablation}), while still outperforming all existing RL baselines (Table~\ref{tab:main_results}), confirming that the gains mainly stem from the reconciled skill-policy co-evolution mechanism rather than the creator's capacity.

\pp{Ablation Study}
Table~\ref{tab:ablation} isolates the contribution of each remaining component. \textit{(1)~Co-evolution order}: replacing simultaneous co-evolution with sequential alternatives (skill-first or policy-first) drops performance by 10.4--18.5\% on \texttt{ALFWorld} and 1.7--2.1\% on \texttt{Search}, showcasing the importance of joint co-evolution rather than separate stages. \textit{(2)~Adaptive weighting}: replacing the recency discount (Eq.~\ref{eq:beta_update}) with last-step only ($w_t{=}0$) drops by 12.9\%/0.9\%, as a single step provides too few samples for reliable skill version selection. Uniform weighting ($w_t{=}1$) narrows the gap to 3.3\%/0.6\% but does not account for policy drift, treating stale evidence equally with recent observations even as the policy evolves. \textit{(3)~Assertion-based analyzer}: removing it in favor of direct per-episode LLM diagnosis costs 5.1\%/0.9\%, showing that a general failure profile across the reservoir yields more targeted skill revisions than individual trajectory inspection.

\begin{wrapfigure}{r}{0.5\textwidth}
\vspace{-12pt}
\centering
\includegraphics[width=0.5\textwidth]{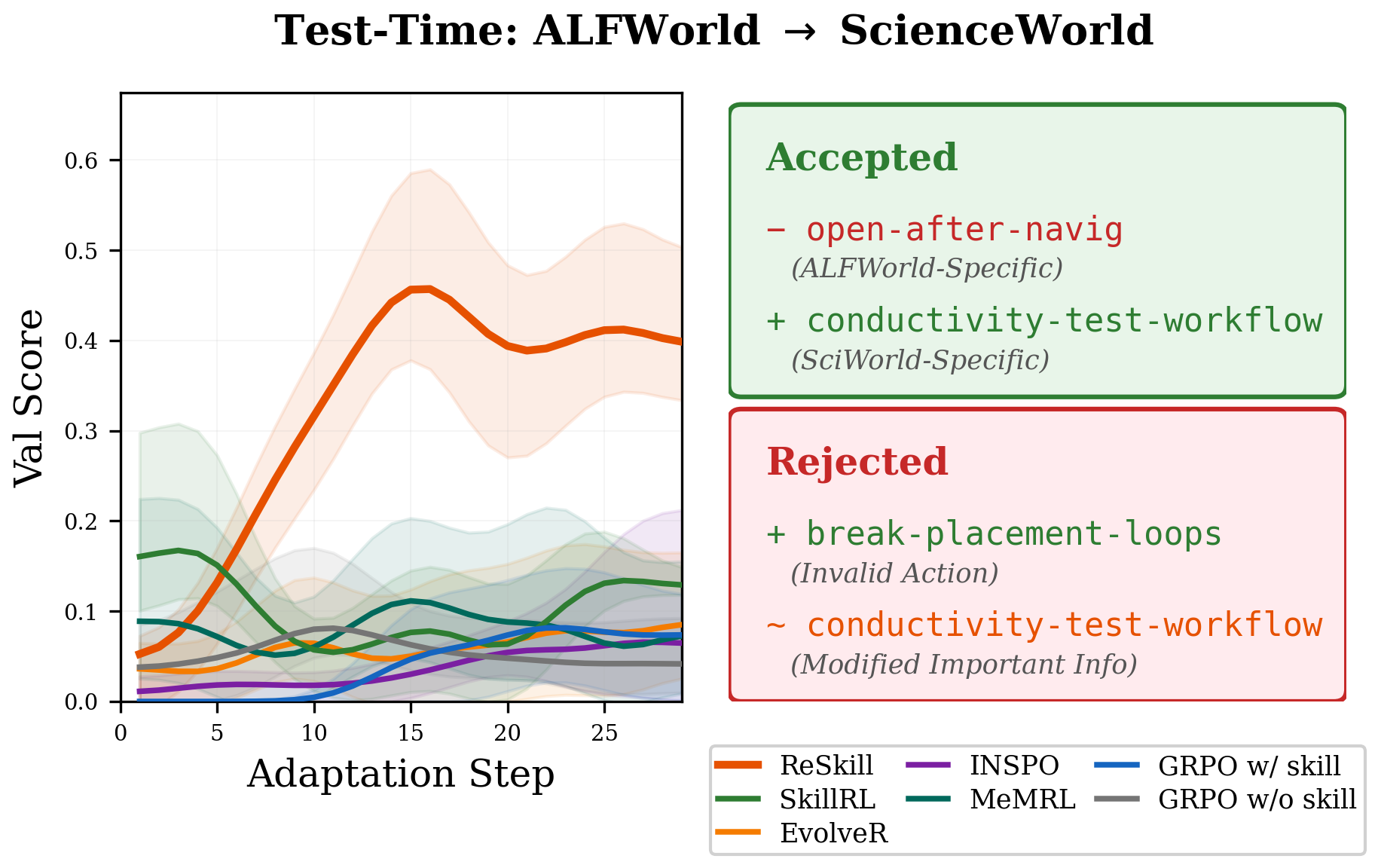}
\caption{Test-time cross-domain adaptation from \texttt{ALFWorld} to \texttt{ScienceWorld}. \ours~rapidly adapts skills to a new domain while baselines remain near zero. Right panel shows accepted and rejected skill operations during adaptation.}
\label{fig:cross_domain_tta}
\vspace{-12pt}
\end{wrapfigure}

\begin{figure*}[!t]
\centering
\includegraphics[width=\textwidth]{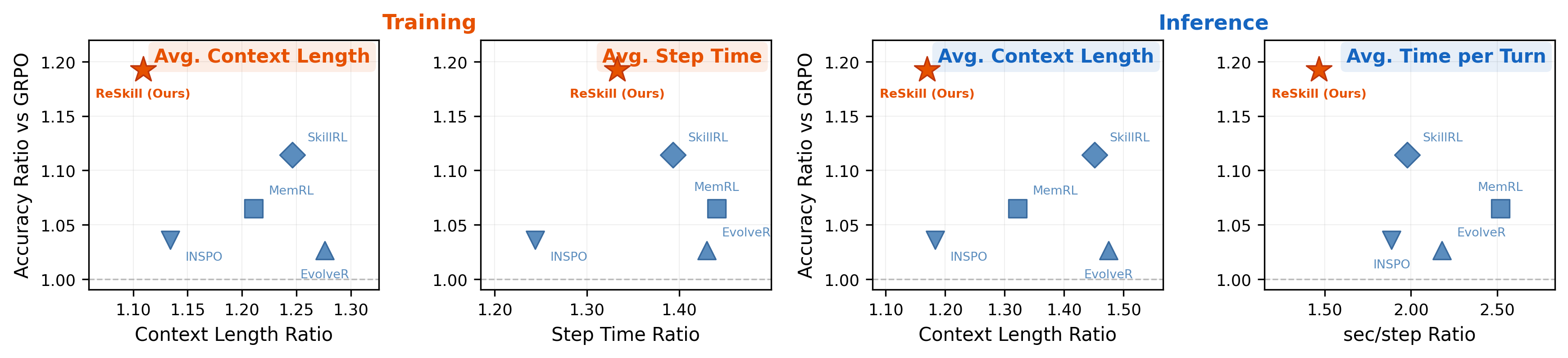}
\caption{Cost analysis on \texttt{ALFWorld} (4B), reported as ratios vs.\ \textsc{Grpo}. \textbf{Left two}: training step time and context length. \textbf{Right two}: inference time per episode and context length. \ours~(star) achieves the highest accuracy ratio while maintaining competitive overhead.}
\label{fig:cost_analysis}
\end{figure*}

\pp{Efficiency}
Figure~\ref{fig:cost_analysis} reports training and inference overhead on \texttt{ALFWorld} (4B) as ratios relative to \textsc{Grpo}. \ours~maintains competitive speed in both training and inference while achieving substantially higher accuracy. Unlike \textsc{Skillrl} and \textsc{Evolver} which load all skills unconditionally across turns, \ours's conditional trigger mechanism fires only relevant skills, keeping context compact over long interactions. As an ablation, we replace conditional triggers with general triggers and observe training context length increase to 1.29$\times$ \textsc{Grpo}, confirming the effectiveness of conditional triggers.

\pp{Test-Time Cross-Domain Adaptation}
To test the generalization capacity of our reconciled training, we freeze the policy trained on \texttt{ALFWorld} and run the full \ours~pipeline at test time on \texttt{ScienceWorld} without policy updates (Figure~\ref{fig:cross_domain_tta}). \ours~rapidly adapts its skill bank to the new domain and achieves strong generalization, while baseline models fail to evolve for the new domain at test time.

\subsection{Qualitative Analysis}

\begin{figure*}[!t]
\centering
\includegraphics[width=\textwidth]{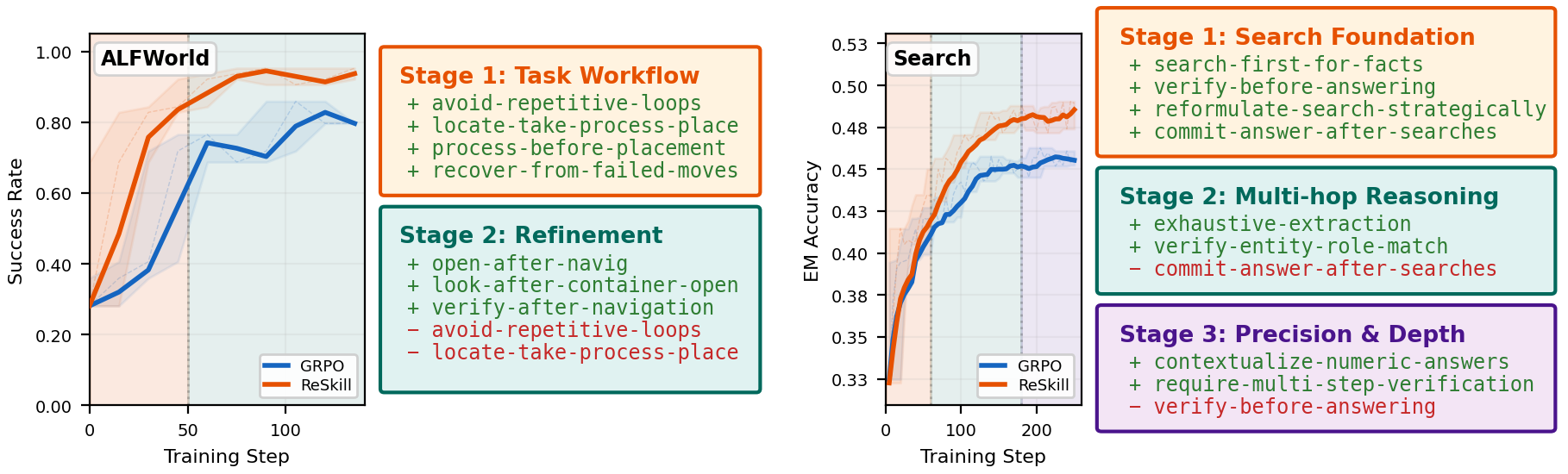}
\caption{Skill-policy co-evolution on \texttt{ALFWorld} (left) and \texttt{Search} (right). Training curves are shown with colored stage bands alongside key skill operations: \textcolor[HTML]{2E7D32}{+}~add (new skill created) and \textcolor[HTML]{C62828}{$-$}~delete (internalized skill pruned).}
\label{fig:skill_lifecycle}
\end{figure*}

\pp{Skill Lifecycle}
Figure~\ref{fig:skill_lifecycle} annotates key skill operations alongside training curves. Both benchmarks exhibit staged co-evolution: early stages build foundational skills (workflow sequencing on \texttt{ALFWorld}, search-then-verify on \texttt{Search}), while later stages prune internalized ones and shift toward refinement and precision as the policy exposes subtler failure modes.

\pp{Case Studies}
We conduct case studies and observe that the model learns to selectively adopt triggered skills as reusable problem-solving strategies within its reasoning process. Details are in Appendix~\ref{app:case_studies}.

\section{Related Work}
\label{sec:related}

We discuss the most closely related work here; broader context on reinforcement learning for LLMs, tool-integrated agents, and bandit methods is provided in Appendix~\ref{app:extended_related}.

\sparagraph{Prompt and Skill Optimization}
The performance of LLM agents is highly sensitive to their guiding instructions \citep{zhou2025multi}. Automated prompt optimization approaches use paraphrasing \citep{zhou2023large}, LLM-based optimizers \citep{yang2024large}, or textual gradients \citep{pryzant-etal-2023-automatic, 2025textgrad} to improve instructions, but operate offline. In the skill domain, prior work extracts experiential rules from trajectories \citep{shinn2023reflexion, zhao2024expel}, builds open-ended skill libraries \citep{wang2023voyager}, and distills reusable workflows from agent experience \citep{wang2024awm}. Anthropic's Skill Creator \citep{anthropic2025skillcreator} automates the creation of modular, trigger-based skills with controlled blind testing, but cannot be directly integrated into a reinforcement learning process. These methods produce valuable skills but do not reconcile creation with policy training.

\pp{Memory and Skill-Based Reinforcement Learning}
Recent work brings skill and memory optimization into RL training: \textsc{Memrl} \citep{zhang2026memrl} evolves episodic memory alongside policy training, \textsc{Evolver} \citep{wu2025evolver} self-distills experience principles retrieved during RL, \textsc{Skillrl} \citep{xia2025skillrl} distills teacher-model skills with trigger-based activation, and \textsc{Inspo} \citep{zhou2025inspo} maintains instruction candidates sampled per rollout group. However, all these approaches treat skill creation and policy learning as separate processes. \ours~closes this gap by reconciling skill evolution with policy training through within-group testing and reward-based selection.


\section{Conclusion}
\label{sec:conclusion}

We introduced \ours, a framework that reconciles skill creation with policy learning by embedding assertion-driven skill revision, within-group skill testing, and adaptive Thompson Sampling into the \textsc{Grpo} training loop at zero additional rollout cost. Across five benchmark domains, \ours~consistently outperforms existing skill and memory-based RL methods, with the largest gains on unseen tasks. Limitations and future directions are discussed in Appendix~\ref{app:limitations}.

\bibliography{refs}
\bibliographystyle{plainnat}

\appendix
\section{Limitations and Future Work}
\label{app:limitations}

\sparagraph{Evolution frequency} As shown in the sensitivity analysis (Figure~\ref{fig:sensitivity}, Appendix~\ref{app:sensitivity}), performance on \texttt{ALFWorld} degrades when evolution frequency is too low ($K_e{=}20$). We observe that infrequent evolution causes the policy to adapt to a fixed skill context; subsequent skill updates then introduce distribution shift that leads to training instability. The evolution frequency remains a hyperparameter requiring per-domain tuning, and we view adaptive evolution scheduling as a promising future direction.

\sparagraph{Skill invocation} Our trigger-based design loads skills into context whenever trigger conditions match, which avoids additional inference steps but gives the model less control over when skills are applied. A natural alternative is to formulate skills as explicit tool calls, providing fine-grained invocation control at the cost of extra generation steps per decision point. Exploring tool-call formulations that maintain rollout efficiency, for example through batched or amortized skill invocation, is a promising direction.

\sparagraph{Skill capacity burden} Adding skills to the context increases the cognitive load on the base model, and we observe two failure modes in practice (see case studies in Appendix~\ref{app:case_studies}). First, smaller models sometimes conflate skill guidance with environment actions, emitting skill names as if they were executable commands rather than incorporating the skill's advice into their reasoning. Second, models may over-apply a skill in situations where it is inapplicable, lacking the capacity to recognize when to disengage. Both failures suggest that skill-augmented RL would benefit from training the policy to treat skills as reasoning-time guidance rather than action-time directives, for example through format-aware reward shaping or explicit separation of skill invocation from action generation.

\section{Experimental Details}
\label{app:experimental_details}

\subsection{Training Details and Hyperparameters}
\label{app:implementation}

We build on the VeRL framework \citep{sheng2025hybridflow} with Ray-based distributed training.

Table~\ref{tab:hyperparams} lists the key training hyperparameters for each benchmark. All benchmarks use the same learning rate and KL loss type; differences arise from environment characteristics (e.g., longer contexts for \texttt{ScienceWorld}, larger batch sizes for \texttt{Search} due to shorter episodes).

\begin{table}[h]
\centering
\caption{Training hyperparameters across benchmarks. All experiments use the Adam optimizer with no warmup. Each node has 8$\times$A100 GPUs.}
\label{tab:hyperparams}
\small
\begin{tabular*}{\textwidth}{@{\extracolsep{\fill}}lccccc}
\toprule
\textbf{Parameter} & \textbf{ALFWorld} & \textbf{Search} & \textbf{ScienceWorld} & \textbf{InterCode-SQL} & \textbf{WANDS} \\
\midrule
\multicolumn{6}{l}{\textit{Reinforcement Learning}} \\
Learning rate & 1e-6 & 1e-6 & 1e-6 & 1e-6 & 1e-6 \\
Batch size & 16 & 256 & 16 & 128 & 16 \\
Mini batch size & 64 & 256 & 64 & 128 & 64 \\
Group size $G$ & 8 & 5 & 8 & 5 & 8 \\
KL coefficient & 0.01 & 0.001 & 0.01 & 0.001 & 0.01 \\
Max prompt length & 4096 & 4096 & 6000 & 4096 & 4096 \\
Max response length & 512 & 512 & 1024 & 512 & 512 \\
Training steps & 120 & 250 & 150 & 150 & 150 \\
Nodes & 2 & 1 & 2 & 2 & 2 \\
\midrule
\multicolumn{6}{l}{\textit{Skill Creator}} \\
Skill bank size & 8 & 8 & 8 & 8 & 8 \\
Evolution frequency $K_e$ & 5 & 5 & 5 & 5 & 5 \\
Exploration bound $\varepsilon_{\text{ts}}$ & 0.15 & 0.15 & 0.15 & 0.15 & 0.15 \\
Reservoir size & 200 & 200 & 200 & 200 & 200 \\
\bottomrule
\end{tabular*}
\end{table}

\subsection{Baseline Implementation Details}
\label{app:baselines}

Unless otherwise specified, all baselines use the original publicly available codebase from their respective papers and share the same training/evaluation data splits as \ours. For RL-based baselines, we use identical environment configurations, learning rates, batch sizes, sequence lengths, and hardware as \ours~(Table~\ref{tab:hyperparams}). Unless otherwise specified, all methods that require an LLM for skill, memory, or instruction creation use the same Claude Sonnet 4.5 API for fair comparison.

\pp{Base Models}
We evaluate two reference models without any training or augmentation. \textit{ReAct} \citep{yao2023react} prompts the base model with a reasoning-plus-action format, interleaving chain-of-thought traces with environment actions. \textit{Claude Sonnet 4.5} uses the same ReAct prompt under identical evaluation protocols, serving as a proprietary reference.

\pp{Evolvement Methods}
\textsc{Reme} \citep{cao2025reme} is a dynamic procedural memory framework that extracts, stores, and retrieves actionable memories from agent trajectories to iteratively improve performance.

\textsc{Memrl} \citep{zhang2026memrl} designs a two-phase retrieval mechanism combining semantic matching with Q-value-driven filtering to identify high-utility memories from past experience. In the evolve-only variant, memories are created from base model rollouts and retrieved at inference without policy updates. We additionally construct a \textsc{Memrl}~(+RL) variant where the same memory system operates concurrently with GRPO policy optimization: Q-values for memory utility are updated online from training rewards while the policy is simultaneously trained, serving as an RL-based baseline. The best of \textsc{Reme} and \textsc{Memrl} is reported as ``Best$^\ddagger$'' in Table~\ref{tab:main_results}.

\pp{Skill-Only Baselines}
We use the high-quality skill bank produced by \textsc{Skillrl}'s~\citep{xia2025skillrl} distillation pipeline (originally created using o3) and directly adopt it as a fixed skill bank to test retrieval-based skill utilization on the base model without RL. We evaluate three retrieval strategies: \texttt{BM25} keyword matching, \texttt{HippoRAG}~\citep{gutierrez2024hipporag} which augments retrieval with knowledge-graph traversal for more semantically grounded skill selection, and long-context loading where the entire skill bank is placed in the model's context window. The best result across these strategies is reported as ``Best$^\dagger$'' in Table~\ref{tab:main_results}.

\pp{RL-Based Methods}
\textsc{Evolver}~\citep{wu2025evolver} operates in alternating cycles of online interaction and offline self-distillation. After each training phase, a distiller LLM analyzes successful and failed trajectories and extracts ``guiding principles'' (from successes) and ``cautionary principles'' (from failures), stored in a curated experience base with semantic deduplication. During training, the agent retrieves relevant principles via embedding similarity to guide its reasoning. Principle quality is tracked via Laplace-smoothed success rates, with low-scoring principles pruned periodically.

\textsc{Inspo}~\citep{zhou2025inspo} maintains a population of instruction candidates sampled per rollout group during GRPO training. Each instruction carries an importance weight updated as a moving average of recent rewards. The population is periodically pruned and replenished by an LLM-based optimizer that reflects on failure trajectories and proposes new candidates. At inference, only the highest-weight instruction is used.

\textsc{Skillrl}~\citep{xia2025skillrl} distills modular skills from training trajectories using a teacher LLM, where each skill consists of a title, an actionable principle, and a triggering condition. Skills are injected into the policy's context via template-based retrieval during both training and evaluation. After validation epochs, failed trajectories trigger recursive skill addition: the teacher identifies unaddressed failure patterns and proposes new skills, which are permanently added to the bank.

\subsection{Benchmark Details}
\label{app:benchmarks}

\pp{ALFWorld}
\texttt{ALFWorld} \citep{ALFWorld20} is an embodied reasoning benchmark where agents interact with household environments via text commands. The action space consists of approximately 14 verb templates (e.g., \texttt{go to}, \texttt{take}, \texttt{put}, \texttt{open}, \texttt{examine}), with a maximum horizon of 50 steps and binary reward (10 on task completion, 0 otherwise). We train on approximately 3,800 games and group the six task types into three categories: Pick\&Place, Transform (averaging Heat, Cool, and Clean), and Examine. Evaluation uses 140 seen and 134 unseen game configurations; we report per-category success rates, seen/unseen averages (micro-averaged over all games in each split), and an overall success rate weighted by split size.

\pp{Agentic Search}
We follow the multi-dataset evaluation protocol of Search-R1 \citep{jin2025searchr1}. Agents answer factual questions via iterative retrieval: the action space is two actions (issue a search query or submit a final answer), with a maximum of 4 turns and binary reward (1 on exact match, 0 otherwise). Retrieval uses E5-base-v2 dense embeddings over a Wikipedia corpus ($\sim$18M passages), returning the top-3 passages per query. We train on $\sim$3,000 questions from \texttt{NQ} \citep{kwiatkowski2019natural} and \texttt{HotpotQA} \citep{yang2018hotpotqa}. Seen evaluation datasets: \texttt{NQ} (300) and \texttt{HotpotQA} (300). Unseen datasets: \texttt{PopQA} \citep{mallen2023popqa} (303), \texttt{TriviaQA} \citep{joshi2017triviaqa} (300), \texttt{2WikiMultiHopQA} \citep{ho2020constructing} (300), \texttt{MuSiQue} \citep{trivedi2022musique} (300), and \texttt{Bamboogle} \citep{press2023bamboogle} (125). We report per-dataset accuracy, seen/unseen averages weighted by dataset size, and overall accuracy across all 1,928 examples.

\pp{WANDS (Web Shopping)}
We construct a web shopping environment inspired by WebShop \citep{yao2022webshop}, built from the WANDS product search dataset \citep{chen2022wands} due to licensing concerns with the original WebShop data. Agents must find and purchase a product matching a goal specification (category, attributes, configurable options, price constraint). The action space is two actions (search or click), with a maximum of 15 steps and binary reward (10 if the correct product is purchased with all required options selected, 0 otherwise). Tasks are categorized by the number of options that must be correctly selected: 2-opt (two required options) and 3-opt (three required options, harder). Training uses 3,000 products from the Furniture domain; evaluation uses 496 held-out products spanning both seen (Furniture) and unseen (Kitchen \& Tabletop) domains. We report accuracy stratified by option complexity (2-opt and 3-opt).

\pp{ScienceWorld}
\texttt{ScienceWorld} \citep{wang2022scienceworld} is an interactive science simulation where agents perform multi-step experiments. We use the electricity split, which contains two task types: identifying power components and testing conductivity. The action space consists of approximately 26 verb templates (e.g., \texttt{pick up}, \texttt{pour}, \texttt{activate}, \texttt{connect}, \texttt{focus on}), with a maximum of 50 steps and binary reward (10 if task score exceeds 70\%, 0 otherwise). For in-distribution evaluation (Test-ID), we use the dev split of 230 episodes with the same task types but held-out variations; for out-of-distribution evaluation (Test-OOD), we use 155 episodes from related but distinct task variants (conductivity with unknown materials, renewable energy). We report ID and OOD accuracy.

\pp{InterCode-SQL}
\texttt{InterCode-SQL} \citep{yang2023intercode} evaluates interactive SQL generation over Spider databases \citep{yu2018spider}. Agents iteratively write and execute SQL queries with execution feedback before submitting a final answer; the action space is two actions (execute exploratory query or submit), with a maximum of 10 turns and binary reward (10 on execution result match with the ground truth, 0 otherwise). Following SQL-Trail \citep{hua2026sql}, training uses 1,027 difficulty-balanced queries, filtered via pass@6 to exclude trivially solvable and impossible cases. Evaluation uses 1,000 examples from 20 held-out Spider databases.

\section{Additional Details and Results}
\label{app:method_details}

\subsection{Pseudocode}

\begin{algorithm}[h]
\begin{footnotesize}
\caption{\ours: Reconciled Skill-Policy Co-Evolution}
\label{alg:rerl}
\begin{algorithmic}[1]
    \STATE \textbf{Input}: Policy $\pi_{\theta_0}$, training steps $T$, evolution frequency $K_e$.
    \STATE \textbf{Initialize}: Skill bank $\mathcal{S} {=} \emptyset$, reservoir $\mathcal{R} {\leftarrow} \emptyset$, posteriors $(\alpha_v, \beta_v) {\leftarrow} (1,1)$ for $v {\in} \{\text{new}, \text{old}\}$, memory $\hat{M} {\leftarrow} \infty$.
    \FOR{$t=1, \dots, T$}
        \STATE \textcolor{paleblue}{[\textbf{Within-Group Skill Testing} (\S\ref{sec:design})]}
        \STATE Sample task $q \sim \mathcal{D}$. Compute $\pi_t(\text{new})$ from the current Beta posteriors; for each rollout $i$, sample $v_i \sim \operatorname{Categorical}(\{\text{new}: \pi_t(\text{new}), \text{old}: 1-\pi_t(\text{new})\})$.
        \STATE Load active skills from $\mathcal{S}_{v_i}$; generate trajectories $\{\tau_i\}_{i=1}^G \sim \pi_{\theta_{t-1}}(\cdot | q, \mathcal{S}_{v_i}; \mathcal{T})$; compute rewards $\{r_i\}$.
        \STATE Update policy: $\theta_t \leftarrow \arg\max_\theta \mathcal{J}_{\text{\ours}}(\theta;\, \theta_{t-1}, \{\tau_i\}, \{r_i\})$. \hfill {\color{gray}\textit{// role 1: policy update}}
        \STATE Add $\{(\tau_i, v_i, r_i)\}$ to experience reservoir $\mathcal{R}$. \hfill {\color{gray}\textit{// role 2: experience collection}}
        \STATE $w_t {\leftarrow} (1{+}n_t/\hat{M})^{-1}$;\; $\alpha_v {\leftarrow} w_t \alpha_v {+} m_t$;\; $\beta_v {\leftarrow} w_t \beta_v {+} (n_t {-} m_t)$,\; for $v {\in} \{\text{new}, \text{old}\}$. \hfill {\color{gray}\textit{// role 3: version evaluation}}
        \IF{$t \bmod K_e = 0$}
            \STATE \textcolor{palered}{[\textbf{Skill Evolution} (\S\ref{sec:skill_evolution})]} Accept $\mathcal{S}_{\text{new}}$ if $\mathbb{E}[p_{\text{new}}] {>} \mathbb{E}[p_{\text{old}}]$; update $\hat{M}$ via predictive-likelihood.
            \STATE \textcolor{palegreen}{[\textbf{Skill Creation} (\S\ref{sec:skill_creator})]} Grade $\mathcal{R}$ with assertions; diagnose; propose new $\mathcal{S}_{\text{new}}$.
            \STATE Reset posteriors: $(\alpha_v, \beta_v) \leftarrow (1, 1)$; begin testing $\mathcal{S}_{\text{new}}$ vs.\ $\mathcal{S}_{\text{old}}$.
        \ENDIF
    \ENDFOR
    \STATE \textbf{Return} optimized policy $\pi_{\theta_T}$ and skill bank $\mathcal{S}$.
\end{algorithmic}
\end{footnotesize}
\end{algorithm}

\subsection{\textsc{Grpo} Objective}
\label{app:grpo}

For completeness, the full \textsc{Grpo} objective \citep{shao2024deepseekmath} is:
\begin{equation}
\begin{aligned}
&\mathcal{J}_{\text{GRPO}}(\theta) = \mathbb{E}_{\tau \sim \pi_{\text{old}}} \Bigg[ \frac{1}{G} \sum_{i=1}^G \frac{1}{\sum_t \mathcal{I}_{i,t}} \sum_{t=1}^{|\tau_i|}\mathcal{I}_{i,t} \cdot \min \left( r_{i,t}\hat{A}_{i}, \text{clip}(r_{i,t}, 1{-}\epsilon, 1{+}\epsilon) \hat{A}_{i} \right)  - \beta \mathbb{D}_{\text{KL}} \Bigg],
\end{aligned}
\label{eq:grpo}
\end{equation}
where
\begingroup
\small
\[
r_{i,t}(\theta) =
\frac{\pi_\theta(\tau_{i,t}|\tau_{i,<t})}
{\pi_{\theta_{\text{old}}}(\tau_{i,t}|\tau_{i,<t})},
\qquad
\hat{A}_i =
\frac{r_i - \text{mean}(\{r_j\}_{j=1}^{G})}
{\text{std}(\{r_j\}_{j=1}^{G})}.
\]
\endgroup
Here, $r_{i,t}(\theta)$ is the token-level importance sampling ratio, $\hat{A}_i$ is the group-normalized advantage, $\epsilon$ is the clip ratio, and $\beta$ controls the KL divergence penalty $\mathbb{D}_{\text{KL}}(\pi_\theta \| \pi_{\text{ref}})$. The indicator $\mathcal{I}_{i,t}$ masks out environment observation tokens so that gradients are computed only over the agent's own tokens \citep{jin2025searchr1, xue2025simpletir}.

\subsection{Skill Creator Pipeline: Full Details}
\label{app:skill_pipeline}

Inspired by Anthropic's Skill Creator \citep{anthropic2025skillcreator}, our skill creator pipeline decomposes skill creation into the following stages: reservoir collection, assertion grading, contrastive analysis, insight aggregation and semantic grouping, skill recommendation, and skill authoring. We present each stage in execution order.

\subsubsection{Reservoir Collection and Contrastive Analysis}

Each \textsc{Grpo} rollout produces trajectories tagged with a version assignment $v_i \in \{\mathcal{S}_{\text{new}}, \mathcal{S}_{\text{old}}\}$ and binary reward $r(\tau_i)$. The reservoir $\mathcal{R}$ stores these trajectories in a last-in-first-out structure, enforcing diversity by retaining at most one success and one failure per version per task. That is, for each task the reservoir holds up to four episodes (new-success, new-failure, old-success, old-failure), replacing older entries as new rollouts arrive.

When skill creation is triggered, the reservoir is sampled into task groups. We prioritize groups with diversity: those containing both successes and failures, and both active and inactive skill conditions. Each group receives a separate LLM call comparing the episodes within it: success vs.\ failure, new-skill vs.\ old-skill, and the specific actions at which behavior diverges. The LLM produces a structured \textit{insight} per group containing: a behavioral finding, a failure-mode label, the failure and success action snippets, an assessment of whether active skills helped or hurt, and a confidence level. This step condenses the raw reservoir into a manageable set of grounded insights for downstream diagnosis.

\subsubsection{Assertion Grading and Diagnosis}

\pp{Assertion Types}
The assertion grader $\mathcal{G}$ maintains a dynamic set of rule-based assertions $\mathcal{A} = \{\phi_j\}$, where each assertion is a predicate over trajectories generated by the LLM-based diagnoser. Assertions capture behavioral properties at varying granularity: whether a particular action pattern appears (or is absent) in a trajectory, whether two actions occur in the correct temporal order (e.g., searching before submitting), whether the agent avoids unproductive repetition, whether the trajectory falls within expected length bounds, and whether the environment's observations indicate the agent reached certain states. The diagnoser creates, modifies, and deletes assertions as the policy's failure landscape evolves.

\pp{Grading}
Given the failure reservoir $\mathcal{R}$, the grader evaluates every trajectory against all assertions, producing per-assertion pass rates:
\begin{equation}
    \hat{r}_j = |\mathcal{R}|^{-1} \sum_{\tau \in \mathcal{R}} \mathbf{1}[\text{check}_j(\tau, \text{params}_j)].
\end{equation}
Here, $\tau$ denotes a trajectory in the reservoir and $\text{params}_j$ denotes assertion-specific parameters. Low pass rates identify prevalent failure modes. The grading is entirely rule-based, enabling evaluation of the full reservoir rather than a subsample.

\pp{Diagnoser}
The diagnoser receives one-liner summaries extracted from the contrastive insights alongside the per-assertion pass rates and the current skill bank summary. It produces three outputs: (1) assertion CRUD operations that evolve $\mathcal{A}$ by adding, deleting, or modifying assertions as behavior shifts; (2) semantic clustering that groups insights by underlying failure mode, selecting a representative per group; and (3) a quantitative failure diagnosis connecting assertion pass rates to the identified clusters.

\subsubsection{Skill Recommendation and Authoring}

\pp{Recommender}
The recommender receives the insight groups (with expanded representative insights showing full action snippets), the current skill library with per-skill diagnostics (trigger rate, age), the version history of prior proposals with their accept/reject outcomes, and the environment's action vocabulary. It proposes 1--3 operations (\textsc{add}, \textsc{modify}, or \textsc{delete}), each specifying: the operation type, intent (what to change and why), and supporting evidence (which insight group motivates it). The recommender does \textit{not} write skill content; it focuses purely on strategic decisions.

The recommender is also informed of the skill lifecycle stage: when the library is below capacity it favors \textsc{add} operations to build coverage, and as skills accumulate toward $K_{\max}$ it shifts toward \textsc{modify} and \textsc{delete} to refine and prune. Beyond lifecycle guidance, it prioritizes unaddressed failure clusters for additions, skills whose guidance contradicts observed successful behavior for modifications, and skills the policy has internalized (chronically low trigger rates or high reward regardless of skill presence) for deletions.

\pp{Author}
The author receives the recommended operations along with selected experience summaries and available action patterns, focusing on creating the skill content.

For each operation, the author writes:
\begin{itemize}[itemsep=2pt, leftmargin=*]
    \item \textbf{Skill content} $c_k$: Three structured fields ($\leq$500 characters total):
    \begin{itemize}[itemsep=1pt]
        \item \texttt{when\_to\_use}: Situational applicability ($\leq$25 words)
        \item \texttt{action}: Concrete behavioral guidance ($\leq$50 words)
        \item \texttt{examples}: DO/DON'T action snippets grounded in the referenced insights
    \end{itemize}
    \item \textbf{Trigger condition} $\text{trig}_k$: Type (general, beginning, or action-pattern) and regex pattern derived from the action vocabulary
\end{itemize}

The author's output constitutes the candidate version $\mathcal{S}_{\text{new}}$, which then enters the trigger validation gate (\S\ref{app:triggers}) before A/B testing.

\subsection{Trigger Condition Types and Validation}
\label{app:triggers}

\subsubsection{Trigger Types}

Each skill $S_k = (c_k, \text{trig}_k)$ has a deterministic trigger condition evaluated at every step of a rollout. The trigger determines whether the skill's content is loaded into the agent's context at that step.

\pp{General Triggers}
Skills with \texttt{trigger\_type = general} fire at every step of every episode. These encode broad strategic principles applicable regardless of the agent's current state.

\pp{Beginning Triggers}
Skills with \texttt{trigger\_type = beginning} fire only at step $t = 0$ (episode initialization). These provide planning-level guidance for how to approach the task without cluttering mid-episode prompts.

\pp{Action-Pattern Triggers}
Skills with \texttt{trigger\_type = action\_pattern} fire when the agent's previous action $a_{t-1}$ matches a regex pattern stored in \texttt{trigger\_pattern}. For example, a skill targeting search refinement might use \texttt{trigger\_pattern = "search\textbackslash[.*\textbackslash]"} to activate only after the agent has just performed a search action. This enables context-sensitive guidance: a skill about interpreting search results only appears after a search, not during unrelated actions.

The trigger evaluation is fully deterministic and requires no LLM calls:
\begin{equation}
    \text{trig}_k(t, a_{t-1}) = \begin{cases}
        1 & \text{if type}_k = \texttt{general} \\
        \mathbf{1}[t = 0] & \text{if type}_k = \texttt{beginning} \\
        \mathbf{1}[\texttt{regex}(\texttt{pattern}_k, a_{t-1})] & \text{if type}_k = \texttt{action\_pattern}
    \end{cases}
\end{equation}

At step $t$, the active skill bank is $\mathcal{S}_{\text{active}}(t) = \{S_k \in \mathcal{S} : \text{trig}_k(t, a_{t-1}) = 1\}$.

\subsubsection{Selective Loading via \texttt{when\_to\_use}}

Even after trigger-based filtering, multiple skills may be active simultaneously. Each skill's content $c_k$ includes a \texttt{when\_to\_use} field describing its situational applicability in natural language. The agent reads these descriptions as part of the prompt and decides during reasoning whether each loaded skill is relevant to the current state. This creates a two-level filtering effect:

\begin{enumerate}[itemsep=2pt]
    \item \textbf{Deterministic trigger}: Gates whether the skill's content enters the prompt at all.
    \item \textbf{In-context applicability}: The \texttt{when\_to\_use} field tells the policy when the loaded guidance should be followed.
\end{enumerate}

The trigger condition serves the routing role, and \texttt{when\_to\_use} serves the relevance-filtering role.

\subsubsection{Trigger Validation and Optimization}

Motivated by Skill Creator's description optimization phase, we validate proposed triggers before the version enters A/B testing to ensure sufficient activation coverage. A skill that never fires cannot be meaningfully evaluated and wastes the testing budget on a version that behaves identically to $\mathcal{S}_{\text{old}}$.

\pp{Trigger Rate Estimation}
For each proposed skill in $\mathcal{S}_{\text{new}}$ with an action-pattern trigger, the pipeline estimates the trigger's firing rate over the reservoir:
\begin{equation}
    \hat{p}_{\text{trigger}} = |\mathcal{R}|^{-1} \sum_{\tau \in \mathcal{R}} \mathbf{1}\!\left[\exists\, t : \texttt{regex}(\texttt{pattern}_k, a_t^{(\tau)}) = \text{true}\right],
\end{equation}
where $a_t^{(\tau)}$ is the action at step $t$ of trajectory $\tau$. This estimates the fraction of episodes in which the skill would activate at least once.

\pp{Validation Gate}
A proposed skill passes validation if $\hat{p}_{\text{trigger}} \geq \delta_{\text{trig}}$ (default $\delta_{\text{trig}} = 0.5$). If any proposed skill in $\mathcal{S}_{\text{new}}$ fails this gate, the entire version proposal is rejected and sent back to the version author for rewriting, with the low trigger rate provided as feedback. The retry loop allows up to 3 attempts before the evolution cycle is skipped entirely. Skills with \texttt{general} or \texttt{beginning} triggers bypass this validation, as they fire unconditionally or at a known rate.

\subsection{Adaptive Staleness Correction: Details}
\label{app:staleness_details}

This appendix motivates the adaptive discount rule $w_t = 1/(1 + n_t/M)$ from Eq.~(\ref{eq:beta_update}), grounds it in the Bayesian posterior update, and provides the full procedure for estimating~$M$ from data.

\subsubsection{Motivation: From Bayesian Updates to Adaptive Discounting}

We begin from the standard Beta-Bernoulli posterior update and show how the discount rule emerges naturally from the desire to control the influence of new observations relative to accumulated evidence.

\pp{The Posterior Mean as a Weighted Average}
After observing $m_t$ successes in $n_t$ episodes at step~$t$, the standard (undiscounted) Beta posterior update $\alpha_t = \alpha_{t-1} + m_t$ yields a posterior mean that is a weighted average of the prior estimate and the new observation:
\begin{equation}
    \mu_t = \frac{S_{t-1}}{S_{t-1} + n_t}\;\mu_{t-1} + \frac{n_t}{S_{t-1} + n_t}\;\frac{m_t}{n_t},
    \label{eq:standard_update}
\end{equation}
where $S_{t-1} = \alpha_{t-1} + \beta_{t-1}$ is the total pseudo-count. The weight on new evidence is $n_t/(S_{t-1} + n_t)$: the more data observed so far ($S_{t-1}$ large), the less each new batch shifts the estimate. In a stationary setting, this is optimal; the posterior concentrates as evidence accumulates.

\pp{The Problem Under Non-Stationarity}
When the policy evolves, the success probability $p_{a,t}$ drifts across steps. Since we evaluate skills alongside an evolving policy, earlier observations reflect a different policy's interaction with the skill and no longer represent the skill's current effectiveness. The accumulated pseudo-count $S_{t-1}$ reflects all past policies, not just the current one. A large $S_{t-1}$ anchors the estimate to stale evidence, making it slow to respond to genuine changes in how the skill performs under the evolving policy. To faithfully assess whether a skill helps the current policy, we must discount this stale evidence so that the accept/reject decision reflects recent, policy-aligned performance.

\pp{Discounting as Effective Sample Size Control}
A geometric discount $\alpha_t = w \cdot \alpha_{t-1} + m_t$ shrinks the accumulated pseudo-count before each update, reducing the effective sample size of old evidence. With discount factor~$w$, the posterior mean becomes:
\begin{equation}
    \mu_t = \frac{w \cdot S_{t-1}}{w \cdot S_{t-1} + n_t}\;\mu_{t-1} + \frac{n_t}{w \cdot S_{t-1} + n_t}\;\frac{m_t}{n_t}.
    \label{eq:discounted_update}
\end{equation}
The key quantity is the ratio of old-to-new influence: $w \cdot S_{t-1} : n_t$. With a constant~$w$, this ratio depends on the entire allocation history through~$S_{t-1}$. In our setting, Thompson Sampling causes the per-arm episode count $n_t$ to vary by 5--10$\times$ across steps. The same constant~$w$ cannot be appropriate for both extremes: after a step with few episodes, the accumulated estimate is noisy and should be retained; after a step with many episodes, the new observation is precise and old evidence should be discounted more aggressively.

\pp{The $M$-Parameterization}
To decouple the discount from the allocation history, we set $w_t = M/(M + n_t) = 1/(1 + n_t/M)$. This adapts automatically to each step's allocation:
\begin{itemize}[itemsep=2pt, leftmargin=*]
    \item When $n_t \ll M$ (few new episodes): $w_t \approx 1$, so old evidence is mostly retained because the new batch is too small to be trusted alone.
    \item When $n_t = M$ (the crossover point): $w_t = 1/2$, so old pseudo-counts are halved and old and new evidence contribute equally.
    \item When $n_t \gg M$ (many new episodes): $w_t \approx 0$, so old evidence is aggressively discounted because the new batch alone provides a reliable estimate.
\end{itemize}
We call $M$ the \emph{memory parameter}: it controls how many recent observations the system effectively remembers. A small $M$ gives short memory (fast adaptation, more noise); a large $M$ gives long memory (slow adaptation, more stability).

\subsubsection{Predictive-Likelihood Criterion}

The memory parameter $M$ need not be hand-tuned: the evolution procedure itself generates the data required to estimate it. Each completed A/B test window produces a sequence of per-step observations $(m_t, n_t)$ for each arm. A well-chosen $M$ should make the discounted posterior \textit{predictive}: given observations from steps $1, \ldots, t{-}1$, the model under a good $M$ should assign high probability to the outcome actually observed at step~$t$. By scoring candidate values of $M$ on how well they predict one step ahead, accumulated across all steps and all completed tests, we obtain a natural maximum-likelihood estimator. Before any test has completed, we initialize with $M \to \infty$ (equivalently $w_t = 1$, no discounting), corresponding to the uninformative case. Once a few test cycles finish, the accumulated one-step-ahead likelihoods provide a reliable signal for $M$.

\pp{Setup}
Suppose we have $K$ completed tests, each spanning $T$ training steps. Test~$k$ provides per-step per-arm data $\mathbf{d}^{(k,a)} = \{(m_t^{(k,a)},\, n_t^{(k,a)})\}_{t=1}^T$ for each arm $a \in \{A, B\}$, where $m_t$ is the number of successes and $n_t$ is the number of episodes assigned to arm~$a$ at step~$t$. We use $\beta_p$ for the Beta parameter to distinguish it from the KL penalty $\beta$ in Eq.~(\ref{eq:lgrpo}).

\pp{BetaBinomial Predictive Probability}
Given current posterior parameters $(\tilde\alpha, \tilde\beta_p)$, the predictive probability of observing $m$ successes in $n$ trials is:
\begin{equation}
    P(m \mid n, \tilde\alpha, \tilde\beta_p) = \binom{n}{m} \frac{B(m + \tilde\alpha,\; n - m + \tilde\beta_p)}{B(\tilde\alpha,\; \tilde\beta_p)},
    \label{eq:betabin}
\end{equation}
where $B(\cdot, \cdot)$ is the Beta function.

\pp{Per-Arm Log-Likelihood}
For a candidate memory parameter $M$ and one arm's data sequence $\mathbf{d} = \{(m_t, n_t)\}_{t=1}^T$, we define the \textit{sequential predictive log-likelihood} $\ell(M;\, \mathbf{d})$ via the following procedure:
\begin{enumerate}[itemsep=2pt]
\item Initialize $\alpha_0 = 1$, $\beta_{p,0} = 1$ \quad (uniform $\mathrm{Beta}(1,1)$ prior).
\item For $t = 1, \ldots, T$:
  \begin{enumerate}
    \item Discount: $\tilde\alpha_t = \frac{M}{M + n_t} \cdot \alpha_{t-1}$, \; $\tilde\beta_{p,t} = \frac{M}{M + n_t} \cdot \beta_{p,t-1}$.
    \item Score: $\ell_t = \log P(m_t \mid n_t, \tilde\alpha_t, \tilde\beta_{p,t})$ \quad (Eq.~\ref{eq:betabin}).
    \item Update: $\alpha_t = \tilde\alpha_t + m_t$, \; $\beta_{p,t} = \tilde\beta_{p,t} + (n_t - m_t)$.
  \end{enumerate}
\item Return $\ell(M;\, \mathbf{d}) = \sum_{t=1}^T \ell_t$.
\end{enumerate}

\pp{Total Objective}
The total log-likelihood over all tests and arms is:
\begin{equation}
    L(M) = \sum_{k=1}^K \sum_{a \in \{A, B\}} \ell\bigl(M;\, \mathbf{d}^{(k,a)}\bigr),
    \label{eq:total_ll}
\end{equation}
and the MLE is $\hat{M} = \arg\max_M L(M)$. Since $M$ is a single scalar and $L(M)$ is cheap to evaluate (a sum of log-likelihoods over completed tests), we find the optimum by grid search over $M \in [1, 2000]$.

\pp{Online Estimation}
Since $L(M)$ decomposes additively across tests, it can be updated incrementally. After each completed test~$k$, we compute $L_k(M) = L_{k-1}(M) + \ell(M;\, \mathbf{d}^{(k,A)}) + \ell(M;\, \mathbf{d}^{(k,B)})$ and re-optimize $\hat{M}_k = \arg\max_M L_k(M)$. This requires $O(|\text{grid}| \times T)$ work per new test, with no need to re-score history.

\subsubsection{Empirical Validation}

\begin{figure*}[t]
\centering
\includegraphics[width=\textwidth]{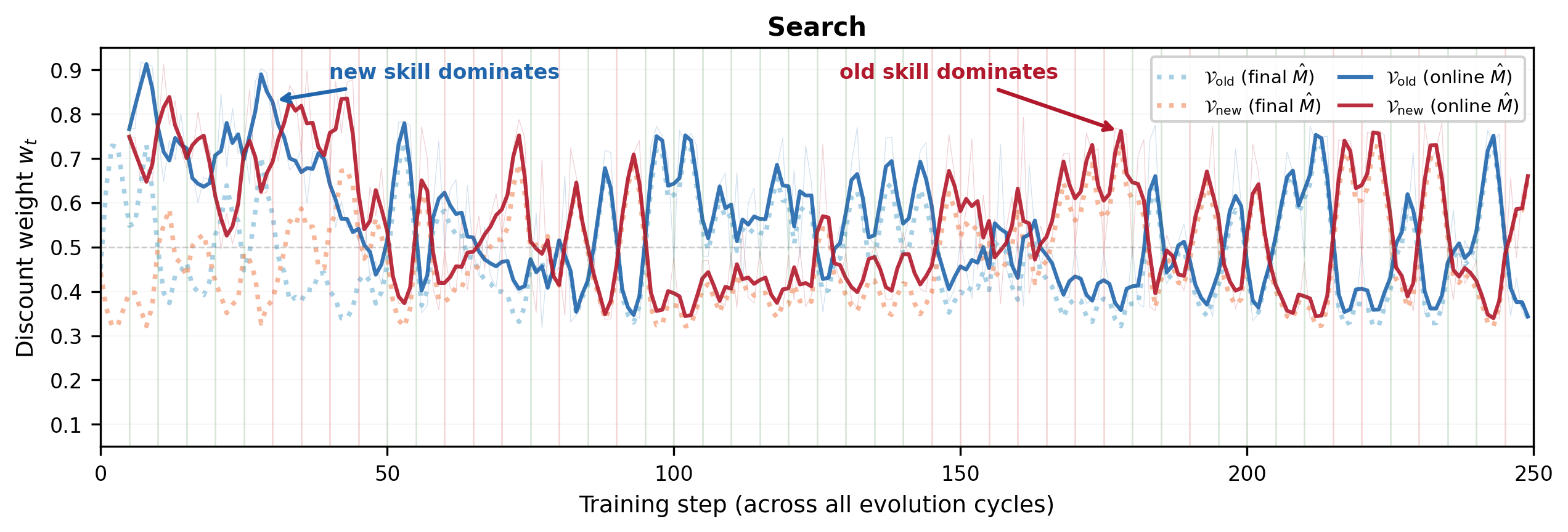}
\caption{Per-arm discount weights $w_t = 1/(1 + n_t/\hat{M})$ during training on \texttt{Search}. Solid lines use the online $\hat{M}_k$; dotted lines use the final $\hat{M}$. The two arms' weights are anti-correlated: Thompson Sampling allocates more episodes to the favored arm, lowering its $w_t$ (more forgetting), while the minority arm retains more history. Green/red vertical lines indicate accepted/rejected evolution cycles.}
\label{fig:adaptive_discount}
\end{figure*}

We validate the online estimation procedure on the \texttt{Search} domain (\cref{fig:adaptive_discount}). Dotted lines show the discount weights under the final $\hat{M}$ (computed from all data up to step 250); solid lines show the weights under the online $\hat{M}_k$ (estimated incrementally after each completed test). During the first ${\sim}$50 steps, the system is still initializing from the uniform prior ($M = \infty$, no discounting) and the online weights gradually deviate from this uninformative starting point as the first test cycles complete. After this warm-up, the online estimate stabilizes and the solid curves closely track the dotted curves, confirming rapid convergence without a separate calibration phase.

The discount weights reveal the adaptive allocation behavior. When Thompson Sampling favors the new skill (allocating more episodes to it), the new arm's $w_t$ drops (more forgetting) while the old arm's $w_t$ rises (more retention), because the old arm receives fewer episodes and its estimate is noisier. This anti-correlation is visible throughout: around step 30, the new skill dominates allocation; conversely, around step 175, the old skill dominates. The discount thus automatically gives more memory to the under-sampled arm at each step.

\subsection{Full Baseline Results}
\label{app:full_baselines}

Table~\ref{tab:main_results} reports per-cell best values for non-RL methods to save space. Table~\ref{tab:full_baselines} expands these into individual method results. All values are mean\tpm{std} over 3 random seeds.

\begin{table*}[h]
\caption{Full results for non-RL baselines (\%). These methods are summarized as Best$^\dagger$ (skill-only) and Best$^\ddagger$ (evolve) in Table~\ref{tab:main_results}. \textsc{BM25}, \textsc{HippoRAG}, and \textsc{Long-context} load skills from \textsc{SkillRL}'s oracle skill bank via retrieval or full injection. \textsc{ReMe} and \textsc{MemRL} evolve memories/procedures without policy training.}
\label{tab:full_baselines}
\centering
\scriptsize
\setlength{\tabcolsep}{4pt}
\textbf{ALFWorld}\\[2pt]
\resizebox{\textwidth}{!}{%
\begin{tabular}{ll cccc @{\hskip 12pt} c @{\hskip 12pt} cccc @{\hskip 18pt} c}
\toprule
& & \multicolumn{4}{c}{Seen} && \multicolumn{4}{c}{Unseen} & \\
\cmidrule(lr){3-6} \cmidrule(lr){8-11}
& Method & Pick\&Place & Transform & Examine & Avg && Pick\&Place & Transform & Examine & Avg & Overall \\
\midrule
\rowcolor{Gray} \multicolumn{12}{c}{\textbf{Qwen3-4B-Instruct}} \\
\midrule
\multirow{3}{*}{\textit{w/ Skill}} & BM25 & 57.1\tpm{4.3} & 26.0\tpm{4.7} & 61.5\tpm{0.0} & 42.4\tpm{4.1} && 43.1\tpm{5.1} & 36.9\tpm{3.1} & 66.7\tpm{5.6} & 42.8\tpm{2.4} & 42.6\tpm{3.1} \\
& HippoRAG & 51.4\tpm{6.8} & 22.5\tpm{2.2} & 53.8\tpm{0.0} & 37.6\tpm{3.9} && 37.4\tpm{5.1} & 43.6\tpm{5.6} & 68.5\tpm{6.4} & 45.0\tpm{4.1} & 41.5\tpm{3.3} \\
& Long-context & 61.6\tpm{2.6} & 43.1\tpm{3.1} & 43.6\tpm{4.4} & 51.0\tpm{2.5} && 50.4\tpm{7.0} & 68.0\tpm{3.5} & 74.1\tpm{8.5} & 63.4\tpm{3.4} & 57.5\tpm{2.6} \\
\midrule
\multirow{2}{*}{\textit{w/ Evolve}} & ReMe & 60.5\tpm{1.6} & 44.6\tpm{0.7} & 79.5\tpm{3.6} & 54.5\tpm{0.9} && 45.5\tpm{3.0} & 58.2\tpm{2.3} & 46.3\tpm{2.6} & 52.7\tpm{0.4} & 53.6\tpm{0.5} \\
& MemRL & 52.0\tpm{2.1} & 30.9\tpm{3.2} & 56.4\tpm{13.1} & 42.1\tpm{2.1} && 41.5\tpm{2.0} & 44.4\tpm{4.4} & 55.6\tpm{7.9} & 45.0\tpm{2.7} & 43.6\tpm{2.2} \\
\midrule
\rowcolor{Gray} \multicolumn{12}{c}{\textbf{Qwen3-8B}} \\
\midrule
\multirow{3}{*}{\textit{w/ Skill}} & BM25 & 79.1\tpm{3.9} & 40.2\tpm{0.8} & 59.0\tpm{4.4} & 58.3\tpm{1.1} && 76.4\tpm{3.7} & 64.0\tpm{1.3} & 68.5\tpm{6.4} & 68.4\tpm{0.4} & 63.6\tpm{0.5} \\
& HippoRAG & 71.2\tpm{2.9} & 40.7\tpm{5.6} & 64.1\tpm{11.8} & 55.7\tpm{1.9} && 55.3\tpm{1.4} & 67.1\tpm{4.1} & 55.6\tpm{11.1} & 61.9\tpm{3.9} & 58.9\tpm{2.4} \\
& Long-context & 78.5\tpm{2.0} & 56.4\tpm{3.1} & 64.1\tpm{4.4} & 66.4\tpm{2.5} && 65.0\tpm{3.7} & 80.4\tpm{1.5} & 72.2\tpm{0.0} & 74.6\tpm{2.0} & 70.7\tpm{1.9} \\
\midrule
\multirow{2}{*}{\textit{w/ Evolve}} & ReMe & 79.1\tpm{4.0} & 58.3\tpm{2.5} & 48.7\tpm{3.6} & 66.2\tpm{2.8} && 53.7\tpm{2.0} & 76.4\tpm{1.7} & 48.1\tpm{2.6} & 65.7\tpm{1.2} & 65.9\tpm{1.8} \\
& MemRL & 64.4\tpm{5.0} & 37.7\tpm{1.4} & 61.5\tpm{0.0} & 51.2\tpm{1.7} && 57.7\tpm{7.0} & 58.7\tpm{2.2} & 57.4\tpm{6.9} & 58.2\tpm{3.8} & 54.9\tpm{2.4} \\
\bottomrule
\end{tabular}}%
\vspace{4pt}

\textbf{Search}\\[2pt]
\resizebox{\textwidth}{!}{%
\begin{tabular}{ll ccc @{\hskip 12pt} c @{\hskip 12pt} cccccc @{\hskip 18pt} c}
\toprule
& & \multicolumn{3}{c}{Seen} && \multicolumn{6}{c}{Unseen} & \\
\cmidrule(lr){3-5} \cmidrule(lr){7-12}
& Method & NQ & HotpotQA & Avg && PopQA & TriviaQA & 2WikiMHQA & MuSiQue & Bamboogle & Avg & Overall \\
\midrule
\rowcolor{Gray} \multicolumn{13}{c}{\textbf{Qwen3-4B-Instruct}} \\
\midrule
\multirow{3}{*}{\textit{w/ Skill}} & BM25 & 29.2\tpm{3.4} & 34.9\tpm{0.2} & 32.1\tpm{1.6} && 39.6\tpm{0.3} & 52.3\tpm{3.2} & 30.9\tpm{2.1} & 9.7\tpm{0.7} & 37.9\tpm{1.7} & 33.6\tpm{0.5} & 33.1\tpm{0.8} \\
& HippoRAG & 31.2\tpm{0.5} & 36.0\tpm{0.3} & 33.6\tpm{0.1} && 42.1\tpm{1.5} & 54.7\tpm{2.4} & 30.1\tpm{0.7} & 10.8\tpm{1.0} & 39.5\tpm{4.0} & 34.9\tpm{1.5} & 34.5\tpm{1.0} \\
& Long-context & 31.6\tpm{0.5} & 35.6\tpm{1.5} & 33.6\tpm{0.9} && 39.9\tpm{1.5} & 55.1\tpm{1.0} & 30.9\tpm{2.4} & 9.3\tpm{0.9} & 42.1\tpm{0.5} & 34.6\tpm{0.6} & 34.3\tpm{0.5} \\
\midrule
\multirow{2}{*}{\textit{w/ Evolve}} & ReMe & 30.4\tpm{1.2} & 35.4\tpm{0.5} & 32.9\tpm{0.8} && 39.4\tpm{0.7} & 57.3\tpm{0.6} & 32.8\tpm{1.5} & 9.0\tpm{1.7} & 40.0\tpm{2.9} & 35.1\tpm{0.7} & 34.5\tpm{0.5} \\
& MemRL & 27.9\tpm{1.0} & 35.6\tpm{1.1} & 31.7\tpm{0.8} && 37.8\tpm{1.2} & 55.1\tpm{0.7} & 34.9\tpm{1.3} & 11.1\tpm{1.0} & 41.3\tpm{2.3} & 35.4\tpm{0.2} & 34.2\tpm{0.2} \\
\midrule
\rowcolor{Gray} \multicolumn{13}{c}{\textbf{Qwen3-8B}} \\
\midrule
\multirow{3}{*}{\textit{w/ Skill}} & BM25 & 32.1\tpm{0.8} & 37.0\tpm{1.5} & 34.6\tpm{0.8} && 39.1\tpm{1.6} & 56.8\tpm{0.7} & 29.8\tpm{0.2} & 11.2\tpm{0.2} & 40.0\tpm{2.4} & 34.8\tpm{0.7} & 34.7\tpm{0.7} \\
& HippoRAG & 31.1\tpm{0.4} & 37.7\tpm{1.2} & 34.4\tpm{0.6} && 42.6\tpm{0.0} & 56.8\tpm{0.8} & 30.6\tpm{1.3} & 10.0\tpm{0.6} & 40.8\tpm{1.6} & 35.5\tpm{0.6} & 35.2\tpm{0.5} \\
& Long-context & 32.8\tpm{0.5} & 39.2\tpm{1.3} & 36.0\tpm{0.4} && 40.6\tpm{1.7} & 57.0\tpm{0.9} & 27.9\tpm{0.7} & 11.1\tpm{1.3} & 40.8\tpm{1.6} & 34.8\tpm{0.3} & 35.2\tpm{0.3} \\
\midrule
\multirow{2}{*}{\textit{w/ Evolve}} & ReMe & 29.5\tpm{0.7} & 36.2\tpm{1.1} & 32.9\tpm{0.5} && 37.5\tpm{0.4} & 57.2\tpm{1.8} & 27.5\tpm{1.7} & 11.9\tpm{1.0} & 40.3\tpm{1.6} & 34.1\tpm{0.6} & 33.8\tpm{0.5} \\
& MemRL & 28.3\tpm{0.3} & 28.3\tpm{1.4} & 28.3\tpm{0.5} && 35.8\tpm{0.7} & 51.9\tpm{1.6} & 24.0\tpm{0.9} & 6.7\tpm{0.7} & 31.7\tpm{0.5} & 29.6\tpm{0.6} & 29.4\tpm{0.9} \\
\bottomrule
\end{tabular}}%
\end{table*}

\subsection{Sensitivity Analysis}
\label{app:sensitivity}

\begin{figure}[h]
\centering
\includegraphics[width=0.5\linewidth]{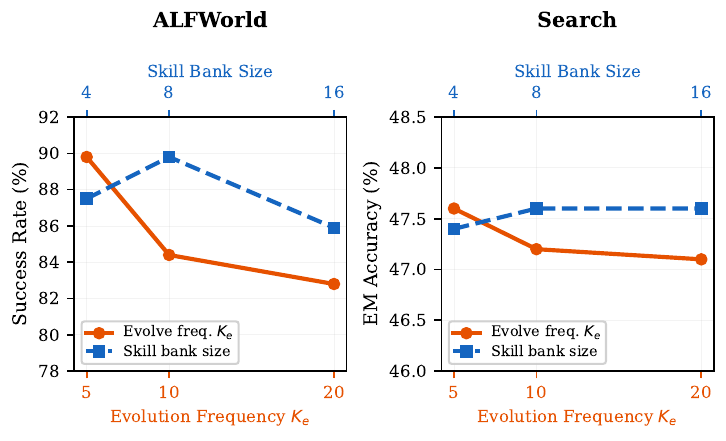}
\caption{Sensitivity to evolution frequency $K_e$ and skill bank size on \texttt{ALFWorld} and \texttt{Search}. Default settings: $K_e{=}5$, bank size 8.}
\label{fig:sensitivity}
\end{figure}

Figure~\ref{fig:sensitivity} examines the effect of evolution frequency $K_e \in \{5, 10, 20\}$ and skill bank size $\{4, 8, 16\}$. On \texttt{ALFWorld} (120 training steps), performance degrades from 89.8\% at $K_e{=}5$ to 87.1\% at $K_e{=}10$ and 85.0\% at $K_e{=}20$. Skill bank size peaks at 8 (89.8\%), with 87.1\% at 4 and 85.9\% at 16. On \texttt{Search} (250 training steps), performance is highly stable across all settings (47.1--47.6\%).

The difference in sensitivity is explained by the ratio of evolution frequency to total training horizon. \texttt{ALFWorld} converges in 120 steps, so $K_e{=}20$ yields only 6 evolution cycles, leaving limited opportunity for skill refinement. \texttt{Search} converges in 250 steps, meaning even $K_e{=}20$ provides 12 evolution cycles, which is sufficient. This suggests that what matters is not the absolute value of $K_e$ but its proportion relative to the training length. Thompson Sampling further stabilizes performance by testing new skill versions against the incumbent before full adoption, preventing premature commitment to harmful revisions.

We consistently use $K_e{=}5$ and bank size 8 across all five experiments without per-domain tuning, demonstrating that these defaults generalize well when the evolution-to-convergence ratio is sufficiently high.

\subsection{Prompt Templates}
\label{app:prompts}

We present the system prompts for each pipeline stage in condensed form. Dynamic content is shown as \texttt{\{placeholder\}}; the full prompts are available in our codebase.

\subsubsection{Contrastive Analyzer}

\begin{templatebox}{Contrastive Analyzer System Prompt}
\small
You analyze a small group of 1--6 episodes that share the same task. Compare success vs failure (and \texttt{new\_skills} vs \texttt{old\_skills} if present) to extract one structured insight.

\vspace{4pt}
Each episode has: Condition (new/old), Outcome (SUCCESS/FAIL), Action trace (\texttt{ACTION:} lines are raw output; \texttt{REASONING:} lines are internal thought), Skills active.

\vspace{4pt}
\texttt{\{episode\_group\}}

\vspace{4pt}
\textbf{Return JSON:}\\
\texttt{\{~"insight": "one-sentence behavioral finding",}\\
\texttt{~~"failure\_mode": "short label for clustering",}\\
\texttt{~~"failure\_snippet": ["action1", "action2", ...],}\\
\texttt{~~"failure\_point": "Step N: what went wrong",}\\
\texttt{~~"success\_snippet": ["action1", "action2", ...],}\\
\texttt{~~"success\_pattern": "what the successful agent did right",}\\
\texttt{~~"skill\_impact": "how existing skills helped or failed",}\\
\texttt{~~"confidence": "high/medium/low"~\}}
\end{templatebox}

\subsubsection{Assertion Diagnoser}

\begin{templatebox}{Assertion Diagnoser System Prompt}
\small
You are a failure analyst for an RL-trained autonomous agent. Your job: (1) evolve assertions (rule-based checks), (2) produce a quantitative failure diagnosis, (3) group insight summaries into semantic clusters.

\vspace{4pt}
\textbf{Assertion Pass Rates}\\
\texttt{\{assertion\_pass\_rates\}}

\vspace{4pt}
\textbf{Insight Summaries}\\
\texttt{\{insight\_summaries\}}

\vspace{4pt}
\textbf{Current Active Skills}\\
\texttt{\{current\_skills\}}

\vspace{4pt}
\textbf{Return JSON:}\\
\texttt{\{~"assertion\_operations": [\{"op": "add/modify/delete", ...~\}],}\\
\texttt{~~"diagnosis": "2--4 sentences connecting pass rates to failures",}\\
\texttt{~~"insight\_groups": [\{"label": "...",}\\
\texttt{~~~~"insight\_indices": [0,3,5], "representative\_index": 0\}]~\}}
\end{templatebox}

\subsubsection{Skill Recommender}

\begin{templatebox}{Skill Recommender System Prompt}
\small
You are a skill library strategist. You receive failure pattern groups (semantically clustered), the current skill library, and version history. Recommend 1--3 skill operations with clear intent and evidence. You do NOT write skill content.

\vspace{4pt}
\textbf{Failure Pattern Groups}\\
\texttt{\{failure\_pattern\_groups\}}

\vspace{4pt}
\textbf{Current Skill Library}\\
\texttt{\{skill\_library\}}

\vspace{4pt}
\textbf{Version History}\\
\texttt{\{version\_history\}}

\vspace{4pt}
\textbf{Environment Action Vocabulary}\\
\texttt{\{action\_vocabulary\}}

\vspace{4pt}
\textbf{Return JSON:}\\
\texttt{\{~"reasoning": "overall strategy (1--2 sentences)",}\\
\texttt{~~"operations": [\{"op": "add/modify/delete",}\\
\texttt{~~~~"target": "skill-name or null",}\\
\texttt{~~~~"intent": "what to change and why",}\\
\texttt{~~~~"evidence": "which insight group supports this",}\\
\texttt{~~~~"referenced\_insights": [0, 3, 7]\}]~\}}
\end{templatebox}

\subsubsection{Skill Author}

\begin{templatebox}{Skill Author System Prompt}
\small
You are a skill content author. Execute each recommended operation by writing precise skill content and trigger conditions.

\vspace{4pt}
\textbf{Recommendations from Skill Recommender}\\
\texttt{\{recommendations\}}

\vspace{4pt}
\textbf{Referenced Insights}\\
\texttt{\{referenced\_insights\}}

\vspace{4pt}
\textbf{Referenced Existing Skills}\\
\texttt{\{referenced\_skills\}}

\vspace{4pt}
\textbf{Environment Action Vocabulary}\\
\texttt{\{action\_vocabulary\}}

\vspace{4pt}
\textbf{Skill content structure} ($\leq$500 characters total):\\
-- \texttt{when\_to\_use}: 1 sentence, $\leq$25 words\\
-- \texttt{action}: 1--2 sentences, $\leq$50 words\\
-- \texttt{examples}: DO/DON'T with actual actions from insights

\vspace{4pt}
\textbf{Return JSON:}\\
\texttt{\{~"operations": [\{"type": "add", "skill": \{"name": "...",}\\
\texttt{~~~~"content": \{"when\_to\_use": "...", "action": "...",}\\
\texttt{~~~~~~"examples": ["DO: ...~$\backslash$n~DON'T: ..."]\},}\\
\texttt{~~~~"trigger\_type": "general/beginning/action\_pattern",}\\
\texttt{~~~~"trigger\_pattern": "regex"\}\}]~\}}
\end{templatebox}

\subsubsection{Agent Runtime Prompt (Skill Injection)}

At each step, triggered skills are injected into the agent's prompt in the following format:

\begin{templatebox}{Agent Prompt with Skill Injection}
\small
\texttt{\{task\_description\}}\\
\texttt{\{memory\_context\}}\\
\texttt{\{triggered\_skills\_section\}}

\vspace{4pt}
\textbf{\#\# Available Skills}

\vspace{2pt}
\textbf{\#\#\# skill: verify-before-submit}\\
~~\texttt{when\_to\_use}: ``Before submitting, verify the answer against retrieved evidence''\\
~~\texttt{action}: ``Re-read the passage and confirm the entity type matches what the question asks for.''\\
~~\texttt{examples}:\\
~~~~-- ``DO: \texttt{<search>} [refined query] \texttt{</search>}''\\
~~~~-- ``DON'T: \texttt{<answer>} [unverified guess] \texttt{</answer>}''

\vspace{4pt}
Now it's your turn to respond for the current step. Check the Available Skills section by reading each \texttt{when\_to\_use} field. If a skill matches your situation, state which skill you are applying and follow its \texttt{action} guidance; otherwise, proceed on your own.

\vspace{2pt}
\texttt{\{action\_format\_instruction\}}
\end{templatebox}

\section{Extended Related Work}
\label{app:extended_related}

\sparagraph{Reinforcement Learning for LLMs}
Reinforcement learning has become a powerful paradigm for post-training LLMs, from alignment with human preferences \citep{ouyang2022training, rafailov2023direct} to reasoning via verifiable rewards. DeepSeek-R1 \citep{guo2025deepseek} demonstrated the effectiveness of RLVR, where \textsc{Grpo} \citep{shao2024deepseekmath} enables group-wise advantage estimation without a critic model. DAPO \citep{yu2025dapo} introduces training stability improvements such as clip-higher and dynamic sampling. Dr.~\textsc{Grpo} \citep{liu2025understanding} rectifies length bias. \ours~is orthogonal to these algorithmic improvements and can be integrated with any \textsc{Grpo}-based training pipeline.

\pp{Tool-Integrated Agents}
LLM agents interact with environments through tool use \citep{yao2023react}. IRCoT \citep{trivedi-etal-2023-interleaving} interleaves chain-of-thought reasoning \citep{wei2022chain} with information retrieval. Toolformer \citep{schick2023toolformer} teaches tool usage via SFT. The RL paradigm has enabled multi-turn tool interactions \citep{feng2025retool, li2025torl}, with search tools excelling at question answering \citep{jin2025searchr1, song2025r1}. End-to-end RL frameworks for tool use \citep{xue2025simpletir, jiang2025verltool} have grown rapidly, yet the role of \textit{dynamic skill guidance} during RL training has been largely unexplored: existing approaches rely on static instructions designed before training begins.

\pp{Bandit Methods for Online Selection}
Bandit algorithms provide a general framework for sequential allocation under uncertainty, balancing exploration and exploitation from partial reward feedback \citep{slivkins2019introduction,lattimore2020bandit}. Thompson Sampling is a posterior-sampling method for this problem \citep{thompson1933likelihood,russo2018tutorial}. Recent LLM systems use bandit formulations for adaptive online decisions such as prompt and context selection \citep{chen2024efficient}, tool or strategy choice \citep{bouneffouf2025surveymultiarmedbanditsmeet}, model routing \citep{dai2024cost,atalar2025neural}, and inference-time personalization \citep{bouneffouf2025surveymultiarmedbanditsmeet}. To our knowledge, there is no direct application of Thompson Sampling to skill evolution inside on-policy agent training, where competing skill versions must be tested using the same rollouts that also update the policy. \ours~fills this gap with within-group version testing and adaptive discounting.

\section{Evolved Skill Examples}
\label{app:evolved_skills}

We present representative skills discovered by \ours~during training.

\subsection{ALFWorld}

\begin{templatebox}{process-before-placement}
\small
\begin{verbatim}
name: process-before-placement
description: "After taking an object requiring heating,
              cooling, or cleaning."
trigger_type: action_pattern
trigger_pattern: "take .+"
\end{verbatim}
\vspace{-4pt}\rule{\linewidth}{0.4pt}\vspace{4pt}

\textbf{when\_to\_use:} After taking an object requiring heating, cooling, or cleaning.

\vspace{3pt}
\textbf{action:} Go directly to the appliance and complete the transformation before navigating to final placement.

\vspace{3pt}
\textbf{examples:}\\
\hspace*{8pt}\textcolor{successgreen}{DO:} \texttt{take bread 1} $\rightarrow$ \texttt{go to fridge 1} $\rightarrow$ \texttt{cool bread 1 with fridge 1} $\rightarrow$ \texttt{go to countertop 1}\\
\hspace*{8pt}\textcolor{failred}{DON'T:} \texttt{take bread 1} $\rightarrow$ \texttt{move bread 1 to countertop 1} (skipped cooling)
\end{templatebox}

\vspace{6pt}
\begin{templatebox}{search-open-surfaces-first}
\small
\begin{verbatim}
name: search-open-surfaces-first
description: "At task start when searching for an object."
trigger_type: beginning
\end{verbatim}
\vspace{-4pt}\rule{\linewidth}{0.4pt}\vspace{4pt}

\textbf{when\_to\_use:} At task start when searching for an object.

\vspace{3pt}
\textbf{action:} Check open surfaces (countertop, diningtable, coffeetable, shelf) before searching closed containers (cabinet, drawer, fridge). Objects are often visible without opening.

\vspace{3pt}
\textbf{examples:}\\
\hspace*{8pt}\textcolor{successgreen}{DO:} \texttt{go to diningtable 1} $\rightarrow$ \texttt{take tomato 2 from diningtable 1}\\
\hspace*{8pt}\textcolor{failred}{DON'T:} \texttt{open cabinet 1} $\rightarrow$ \texttt{open cabinet 2} $\rightarrow$ \ldots{} (15 cabinets without checking surfaces)
\end{templatebox}

\subsection{Search}

\begin{templatebox}{verify-entity-role-match}
\small
\begin{verbatim}
name: verify-entity-role-match
description: "Apply after finding a candidate answer in
              search results."
trigger_type: action_pattern
trigger_pattern: "<search>.*</search>"
\end{verbatim}
\vspace{-4pt}\rule{\linewidth}{0.4pt}\vspace{4pt}

\textbf{when\_to\_use:} Apply after finding a candidate answer in search results.

\vspace{3pt}
\textbf{action:} Verify the entity you extract performs the exact role the question asks about. If question asks ``first governor,'' confirm your entity was first governor, not second. If question asks ``cyclist,'' confirm entity is cyclist, not reporter covering cycling.

\vspace{3pt}
\textbf{examples:}\\
\hspace*{8pt}\textcolor{successgreen}{DO:} Extract ``Ben Foster'' when question asks who played cyclist and results show Foster played Armstrong\\
\hspace*{8pt}\textcolor{failred}{DON'T:} Extract ``Nathan Wiley'' when results show Wiley played reporter, not the cyclist
\end{templatebox}

\vspace{6pt}
\begin{templatebox}{decompose-multi-hop-questions}
\small
\begin{verbatim}
name: decompose-multi-hop-questions
description: "At the start of multi-hop factual questions."
trigger_type: beginning
\end{verbatim}
\vspace{-4pt}\rule{\linewidth}{0.4pt}\vspace{4pt}

\textbf{when\_to\_use:} At the start of multi-hop factual questions.

\vspace{3pt}
\textbf{action:} Before the first search, analyze if the question has nested dependencies. If yes, plan 2--3 sequential searches: (1) identify intermediate entity, (2) search for final property. At every step, verify the answer addresses the ULTIMATE question, not an intermediate finding.

\vspace{3pt}
\textbf{examples:}\\
\hspace*{8pt}\textcolor{successgreen}{DO:} Q: ``when was the show that person X appeared on first aired'' $\rightarrow$ search for show name $\rightarrow$ search for air date\\
\hspace*{8pt}\textcolor{failred}{DON'T:} Answer with the show's name (intermediate hop, not the final answer)
\end{templatebox}

\subsection{ScienceWorld}

\begin{templatebox}{decide-after-three-focuses}
\small
\begin{verbatim}
name: decide-after-three-focuses
description: "After any focus action on target object."
trigger_type: action_pattern
trigger_pattern: "focus on"
\end{verbatim}
\vspace{-4pt}\rule{\linewidth}{0.4pt}\vspace{4pt}

\textbf{when\_to\_use:} After any focus action on target object.

\vspace{3pt}
\textbf{action:} Move object to box now. Metal to conductive box, all else to nonconductive. Never focus again.

\vspace{3pt}
\textbf{examples:}\\
\hspace*{8pt}\textcolor{successgreen}{DO:} \texttt{move metal fork to green box}\\
\hspace*{8pt}\textcolor{failred}{DON'T:} \texttt{focus on metal fork} (continuing to observe without acting)
\end{templatebox}

\vspace{6pt}
\begin{templatebox}{stop-after-two-failed-moves}
\small
\begin{verbatim}
name: stop-after-two-failed-moves
description: "After 4 total failed move attempts to any box."
trigger_type: action_pattern
trigger_pattern: "move"
\end{verbatim}
\vspace{-4pt}\rule{\linewidth}{0.4pt}\vspace{4pt}

\textbf{when\_to\_use:} After 4 total failed move attempts to any box.

\vspace{3pt}
\textbf{action:} Navigate to workshop if not there, then stop all move attempts. Try different approach or end task.

\vspace{3pt}
\textbf{examples:}\\
\hspace*{8pt}\textcolor{successgreen}{DO:} \texttt{go door to workshop}\\
\hspace*{8pt}\textcolor{failred}{DON'T:} \texttt{move glass jar to purple box} (after 4+ failed moves)
\end{templatebox}

\subsection{WANDS}

\begin{templatebox}{commit-after-search}
\small
\begin{verbatim}
name: commit-after-search
description: "After executing a search and viewing results."
trigger_type: action_pattern
trigger_pattern: "search\[.*\]"
\end{verbatim}
\vspace{-4pt}\rule{\linewidth}{0.4pt}\vspace{4pt}

\textbf{when\_to\_use:} After executing a search and viewing results.

\vspace{3pt}
\textbf{action:} Click a matching product from results instead of searching again. If criteria are met, proceed to product page rather than refining the query.

\vspace{3pt}
\textbf{examples:}\\
\hspace*{8pt}\textcolor{successgreen}{DO:} \texttt{search[query]} $\rightarrow$ \texttt{click[w25800]}\\
\hspace*{8pt}\textcolor{failred}{DON'T:} \texttt{search[query]} $\rightarrow$ \texttt{search[similar query]} repeatedly
\end{templatebox}

\vspace{6pt}
\begin{templatebox}{select-options-before-buying}
\small
\begin{verbatim}
name: select-options-before-buying
description: "After viewing a product detail page with
              selectable options."
trigger_type: action_pattern
trigger_pattern: "click\[w\d+\]"
\end{verbatim}
\vspace{-4pt}\rule{\linewidth}{0.4pt}\vspace{4pt}

\textbf{when\_to\_use:} After viewing a product detail page with selectable options.

\vspace{3pt}
\textbf{action:} Select all required product options (color, material, size) before clicking buy now. Check the product page for option dropdowns and select matching values.

\vspace{3pt}
\textbf{examples:}\\
\hspace*{8pt}\textcolor{successgreen}{DO:} \texttt{click[color: red]} $\rightarrow$ \texttt{click[buy now]}\\
\hspace*{8pt}\textcolor{failred}{DON'T:} \texttt{click[buy now]} immediately after \texttt{click[w02505]}
\end{templatebox}

\subsection{InterCode-SQL}

\begin{templatebox}{filter-with-where-and-nulls}
\small
\begin{verbatim}
name: filter-with-where-and-nulls
description: "Apply when filtering records or using
              subqueries."
trigger_type: general
\end{verbatim}
\vspace{-4pt}\rule{\linewidth}{0.4pt}\vspace{4pt}

\textbf{when\_to\_use:} Apply when filtering records or using subqueries.

\vspace{3pt}
\textbf{action:} Always include WHERE clauses for filtering conditions. In subqueries that find maximum values, handle NULL cases by filtering them out before aggregation.

\vspace{3pt}
\textbf{examples:}\\
\hspace*{8pt}\textcolor{successgreen}{DO:} \texttt{SELECT name FROM Scientists WHERE ssn IN (SELECT scientist FROM AssignedTo WHERE project = (SELECT code FROM Projects WHERE hours = (SELECT MAX(hours) FROM Projects)))}\\
\hspace*{8pt}\textcolor{failred}{DON'T:} \texttt{SELECT DISTINCT College FROM match\_season} (missing filter)
\end{templatebox}

\vspace{6pt}
\begin{templatebox}{fuzzy-text-matching}
\small
\begin{verbatim}
name: fuzzy-text-matching
description: "Apply when filtering on descriptive text
              fields like names, descriptions, or outcomes."
trigger_type: general
\end{verbatim}
\vspace{-4pt}\rule{\linewidth}{0.4pt}\vspace{4pt}

\textbf{when\_to\_use:} Apply when filtering on descriptive text fields like names, descriptions, or outcomes.

\vspace{3pt}
\textbf{action:} Use LIKE with wildcards (\texttt{\%term\%}) instead of exact equality. If results are insufficient, expand to semantically related terms with OR clauses.

\vspace{3pt}
\textbf{examples:}\\
\hspace*{8pt}\textcolor{successgreen}{DO:} \texttt{WHERE outcome\_description LIKE '\%published\%'}\\
\hspace*{8pt}\textcolor{failred}{DON'T:} \texttt{WHERE outcome\_description = 'Published'} (exact match fails on case/phrasing)
\end{templatebox}

\section{Case Studies: Skill-Guided Reasoning}
\label{app:case_studies}

We present representative reasoning traces illustrating how evolved skills shape agent behavior.

\subsection{Search}

\vspace{2pt}
\noindent\textcolor{stage1color}{\rule{3pt}{10pt}}\hspace{4pt}\textbf{Factual Lookup}

\vspace{2pt}
\begin{casebox}[stage1color]
\small
\textbf{Q:} \textit{``Who sings let's hear it for the boy?''}

\vspace{3pt}
\textbf{\textcolor{stage1color}{[1]}} Applying \textbf{\textcolor{skillcolor}{search-first-for-facts}}: the question requires specific factual information about a song performer.
\action{<search> who sings lets hear it for the boy </search>}

\vspace{2pt}
\textbf{\textcolor{stage1color}{[2]}} Applying \textbf{\textcolor{skillcolor}{verify-before-answering}}: the results confirm Deniece Williams is the original singer, featured in the Footloose soundtrack and reaching \#1 on the Billboard Hot 100.
\action{<answer> Deniece Williams </answer>}

\vspace{4pt}
\successbadge\ \small This early-stage skill pattern guides basic actions: first retrieve evidence with a targeted search, then verify the evidence before answering.
\end{casebox}

\vspace{6pt}
\noindent\textcolor{stage1color}{\rule{3pt}{10pt}}\hspace{4pt}\textbf{Entity Role Verification}

\vspace{2pt}
\begin{casebox}[stage1color]
\small
\textbf{Q:} \textit{``Are Thurston Moore and Pharrell Williams both rappers?''}

\vspace{3pt}
\textbf{\textcolor{stage1color}{[1]}}
\action{<search> are Thurston Moore and Pharrell Williams rappers </search>}

\vspace{2pt}
\textbf{\textcolor{stage1color}{[2]}} Applying \textbf{\textcolor{skillcolor}{verify-entity-role-match}}: must ensure both are \textbf{explicitly described as rappers}. Results show Thurston Moore is a guitarist (Sonic Youth), \textbf{not a rapper}. Need to verify Pharrell's role separately.
\action{<search> is Pharrell Williams a rapper </search>}

\vspace{2pt}
\textbf{\textcolor{stage1color}{[3]}} Pharrell is explicitly described as rapper/singer/producer. \textbf{Since only one of the two is a rapper}, the answer is no.
\action{<answer> no </answer>}

\vspace{4pt}
\successbadge\ \small The agent checks each entity's \textit{specific role} rather than just confirming name presence.
\end{casebox}

\vspace{6pt}
\noindent\textcolor{stage1color}{\rule{3pt}{10pt}}\hspace{4pt}\textbf{Multi-hop Decomposition}

\vspace{2pt}
\begin{casebox}[stage1color]
\small
\textbf{Q:} \textit{``The founder of `Black Veil Brides' appears in an upcoming American thriller film written by what two people?''}

\vspace{3pt}
\textbf{\textcolor{stage1color}{[1]}}
\action{<search> Black Veil Brides founder upcoming thriller film writers </search>}

\vspace{2pt}
\textbf{\textcolor{stage1color}{[2]}} Applying \textbf{\textcolor{skillcolor}{require-multi-step-verification}}: must ensure the founder, film, and \textbf{both writers} are explicitly identified. Initial results mention ``The Veil (2017)'' but do not connect to Black Veil Brides.
\action{<search> who is the founder of Black Veil Brides </search>}

\vspace{2pt}
\textbf{\textcolor{stage1color}{[3]}} Confirmed Andy Biersack is the founder. No mention of a thriller film yet.
\action{<search> Andy Biersack upcoming American thriller film </search>}

\vspace{2pt}
\textbf{\textcolor{stage1color}{[4]}} Results reveal he appears in ``American Satan,'' written by \textbf{Ash Avildsen and Matty Beckerman}.
\action{<answer> Ash Avildsen and Matty Beckerman </answer>}

\vspace{4pt}
\successbadge\ \small 3-hop decomposition (founder $\to$ film $\to$ writers) with explicit multi-step verification.
\end{casebox}

\subsection{ALFWorld}

\vspace{2pt}
\noindent\textcolor{stage2color}{\rule{3pt}{10pt}}\hspace{4pt}\textbf{Error Recovery}

\vspace{2pt}
\begin{casebox}[stage2color]
\small
\textbf{Task:} \textit{``Find two handtowel and put them in toilet.''}

\vspace{3pt}
\textbf{\textcolor{stage2color}{[13]}} \textit{Obs: ``Nothing happens.''} Applying \textbf{\textcolor{skillcolor}{diagnose-failed-take}}: the take action failed. Verifying environment state before retrying.
\action{look}

\vspace{2pt}
\textbf{\textcolor{stage2color}{[14]}} Environment confirms handtowel~2 is present. Reattempt with verified state.
\action{take handtowel 2 from handtowelholder 2}

\vspace{4pt}
\successbadge\ \small The skill drives a look-verify-retry protocol: instead of blind retry, the agent confirms object state before reattempting.
\end{casebox}

\vspace{6pt}
\noindent\textcolor{stage2color}{\rule{3pt}{10pt}}\hspace{4pt}\textbf{Navigation-Driven Action}

\vspace{2pt}
\begin{casebox}[stage2color]
\small
\textbf{Task:} \textit{``Cool some bowl and put it in countertop.''}

\vspace{3pt}
\textbf{\textcolor{stage2color}{[19]}} Applying \textbf{\textcolor{skillcolor}{acquire-after-navigation}}: arrived at countertop~3 with cooled bowl. Act immediately.
\action{move bowl 1 to countertop 3}

\vspace{4pt}
\successbadge\ \small The skill triggers on navigation completion and teaches the model to immediately act on what it finds at each location.
\end{casebox}

\subsection{SciWorld}

\vspace{2pt}
\noindent\textcolor{stage3color}{\rule{3pt}{10pt}}\hspace{4pt}\textbf{Skill-Driven Object Naming}

\vspace{2pt}
\begin{casebox}[stage3color]
\small
\textbf{Task:} \textit{``Determine if glass jar is electrically conductive. Located around the workshop.''}

\vspace{3pt}
\textbf{\textcolor{stage3color}{[1]}} ``I am applying the skill \textbf{\textcolor{skillcolor}{extract-full-object-name}} to ensure I use the exact name of the object as stated in the task. The object is `glass jar,' so I will use this exact term.''
\action{look at glass jar}

\vspace{2pt}
\textbf{\textcolor{stage3color}{[2]}} ``Glass jar is electrically nonconductive. I will place it in the purple box.''
\action{move glass jar to purple box}

\vspace{4pt}
\successbadge\ \small This skill was created because earlier episodes showed the agent confusing ``glass jar'' with ``glass cup.'' It fires as the very first action, locks in the correct object name, and produces the minimum-possible-step solution. Directly contrasts pre-skill failures where the agent spent 20+ steps manipulating the wrong object.
\end{casebox}

\vspace{6pt}
\noindent\textcolor{stage3color}{\rule{3pt}{10pt}}\hspace{4pt}\textbf{Commitment After Observation}

\vspace{2pt}
\begin{casebox}[stage3color]
\small
\textbf{Task:} \textit{``Determine if aluminum foil is electrically conductive. Located in workshop. Conductive $\to$ orange box. Nonconductive $\to$ yellow box.''}

\vspace{3pt}
\textbf{\textcolor{stage3color}{[1--5]}} Navigation: \texttt{look around} $\to$ \texttt{go kitchen} $\to$ \texttt{look around} $\to$ \texttt{go door to hallway} $\to$ \texttt{go door to workshop}

\vspace{2pt}
\textbf{\textcolor{stage3color}{[6]}} Now in the workshop. The aluminum foil is present. I must first focus on it to determine its properties.
\action{focus on aluminum foil}

\vspace{2pt}
\textbf{\textcolor{stage3color}{[7]}} I am currently focused on the aluminum foil, which is a metal. Metals are known to be electrically conductive. Since this is the first focus, I need to continue focusing to meet the requirement of 2--3 times before deciding. I will apply \textbf{\textcolor{skillcolor}{decide-after-three-focuses}} to make a decision after three focuses.
\action{focus on aluminum foil}

\vspace{2pt}
\textbf{\textcolor{stage3color}{[8]}} Applying \textbf{\textcolor{skillcolor}{decide-after-three-focuses}}: I have focused on the aluminum foil three times (steps 6, 7, 8), and aluminum is a metal known to be electrically conductive. Metals should be placed in the orange box.
\action{move aluminum foil to orange box}

\vspace{4pt}
\successbadge\ \small In this environment, a single focus suffices to determine conductivity, yet the agent hesitates and repeats the safe ``focus'' action. The skill provides a commitment threshold that breaks the observation loop. While the threshold of three focuses is conservative, the model interprets the skill adaptively to avoid repetitive actions and commit to a decision.
\end{casebox}

\vspace{6pt}
\noindent\textcolor{stage3color}{\rule{3pt}{10pt}}\hspace{4pt}\textbf{Recovery via Workshop Navigation}

\vspace{2pt}
\begin{casebox}[stage3color]
\small
\textbf{Task:} \textit{``Determine if metal pot is electrically conductive. Located in kitchen. Conductive $\to$ yellow box. Nonconductive $\to$ purple box.''}

\vspace{3pt}
\textbf{\textcolor{stage3color}{[1--7]}} Navigate to kitchen, find metal pot, focus twice, then attempt placement:
\action{move metal pot to yellow box} --- \textit{Fails (not in workshop)}

\vspace{2pt}
\textbf{\textcolor{stage3color}{[8--20]}} Agent retries the move 10 more times from the kitchen and hallway, all failing. Gets the object with \texttt{get metal pot}, continues attempting moves from wrong locations.

\vspace{2pt}
\textbf{\textcolor{stage3color}{[21]}} Applying \textbf{\textcolor{skillcolor}{stop-after-two-failed-moves}}: multiple failed attempts to move the metal pot to the yellow box. After consecutive failed moves, I should stop all move attempts and \textbf{navigate to the workshop} where the colored boxes are located.
\action{go door to workshop}

\vspace{2pt}
\textbf{\textcolor{stage3color}{[22--23]}} Now in the workshop. The metal pot is a metal object, therefore electrically conductive. Place in yellow box.
\action{get metal pot\\}
\action{move metal pot to yellow box}

\vspace{4pt}
\successbadge\ \small The skill fires after repeated move failures and redirects the agent to the workshop, where all colored boxes are located. This encodes a spatial layout discovery.
\end{casebox}

\vspace{6pt}
\noindent\textcolor{stage3color}{\rule{3pt}{10pt}}\hspace{4pt}\textbf{The Meta-Bug: Skill Name as Environment Action}

\vspace{2pt}
\begin{casebox}[stage3color]
\small
\textbf{Task:} \textit{``Determine if glass jar is electrically conductive. Located around the kitchen.''}

\vspace{3pt}
\textbf{\textcolor{stage3color}{[1--3]}} \texttt{look around} $\to$ \texttt{go hallway} $\to$ \texttt{go door to kitchen}

\vspace{2pt}
\textbf{\textcolor{stage3color}{[4--8]}} \texttt{focus on glass jar containing nothing} $\times$ 5

\vspace{2pt}
\textbf{\textcolor{stage3color}{[9]}} \texttt{decide-after-three-focuses} \hfill (\textbf{INVALID} --- skill name emitted as action!)

\vspace{2pt}
\textbf{\textcolor{stage3color}{[10--13]}} Alternates between \texttt{focus on glass jar} and \texttt{decide-after-three-focuses} (invalid each time).

\vspace{2pt}
\textbf{\textcolor{stage3color}{[14--30]}} Never places object; episode terminates.

\vspace{4pt}
\failbadge\ \small The agent \textit{understands} the skill (correctly identifies when it should fire and what it should cause) but confuses reasoning-level invocation with environment-level action. It outputs the skill name as if it were a command. This meta-bug directly motivated creation of \texttt{no-skill-invocation}, demonstrating how failures drive skill evolution.
\end{casebox}

\subsection{WANDS (Web Shopping)}

\vspace{2pt}
\noindent\textcolor{stage2color}{\rule{3pt}{10pt}}\hspace{4pt}\textbf{Foundational Purchase Workflow}

\vspace{2pt}
\begin{casebox}[stage2color]
\small
\textbf{Task:} \textit{``Find me end table storage end table with color: black, and price lower than 200.00 dollars.''}

\vspace{3pt}
\textbf{\textcolor{stage2color}{[1]}}
\action{search[end table storage color: black price: less than 200.00]}

\vspace{2pt}
\textbf{\textcolor{stage2color}{[2]}} I am applying the skill: \textbf{\textcolor{skillcolor}{commit-after-search}} because a product matching the criteria (black color, price under \$200) is available in the search results. The Evony block end table with storage (W34267) is priced at \$166.04. To avoid further searches, I will proceed with viewing its details.
\action{click[w34267]}

\vspace{2pt}
\textbf{\textcolor{stage2color}{[3]}} I am applying the skill: \textbf{\textcolor{skillcolor}{select-options-before-buying}} because the product page shows selectable options (color and material), and I need to ensure the correct color (black) is selected before proceeding to buy.
\action{click[color: black]}

\vspace{2pt}
\textbf{\textcolor{stage2color}{[4]}} I am applying \textbf{\textcolor{skillcolor}{complete-purchase-workflow}} because the product meets the required criteria and the ``Buy Now'' option is available.
\action{click[buy now]}

\vspace{4pt}
\successbadge\ \small All three foundational skills fire in sequence: commit to a matching product, select required options, then purchase. This 4-step workflow becomes the scaffold that later, more specialized skills refine.
\end{casebox}

\vspace{6pt}
\noindent\textcolor{stage2color}{\rule{3pt}{10pt}}\hspace{4pt}\textbf{Multi-Skill Recovery with Branching}

\vspace{2pt}
\begin{casebox}[stage2color]
\small
\textbf{Task:} \textit{``Find me glossy circle Iittala bowl, white, porcelain china, under \$90.''}

\vspace{3pt}
\textbf{\textcolor{stage2color}{[1]}} \textbf{\textcolor{skillcolor}{limit-search-retries}} fires.
\action{search[Iittala bowl white porcelain]}

\vspace{2pt}
\textbf{\textcolor{stage2color}{[2]}}
\action{click[w21216]}

\vspace{2pt}
\textbf{\textcolor{stage2color}{[3]}} \textbf{\textcolor{skillcolor}{require-color-selection}} fires --- white NOT in available colors.
\action{click[back to search]}

\vspace{2pt}
\textbf{\textcolor{stage2color}{[4]}} \textbf{\textcolor{skillcolor}{no-description-tabs}} prevents tab exploration.
\action{click[w19864]}

\vspace{2pt}
\textbf{\textcolor{stage2color}{[5]}} \textbf{\textcolor{skillcolor}{require-color-selection}} fires --- white IS available.\\
``I am applying the skill: \textbf{\textcolor{skillcolor}{require-color-selection}} because the task specifies `color: white' and the color dropdown is visible.''
\action{click[color: white]}

\vspace{2pt}
\textbf{\textcolor{stage2color}{[6]}} \textbf{\textcolor{skillcolor}{no-back-after-options}} fires.
\action{click[material: porcelain china]}

\vspace{2pt}
\textbf{\textcolor{stage2color}{[7]}} \textbf{\textcolor{skillcolor}{buy-immediately-after-options}} fires.
\action{click[buy now]}

\vspace{4pt}
\successbadge\ \small Five distinct skills compose into a conditional state machine with branching: \texttt{require-color-selection} plays a dual role, blocking premature purchase when color is available but providing the only legitimate escape when color is unavailable.
\end{casebox}

\subsection{InterCode-SQL}

\vspace{2pt}
\noindent\textcolor{stage1color}{\rule{3pt}{10pt}}\hspace{4pt}\textbf{Structured Verification}

\vspace{2pt}
\begin{casebox}[stage1color]
\small
\textbf{Database:} world\_1 \hfill \textbf{Q:} \textit{``Count the number of countries in Asia.''}

\vspace{3pt}
\textbf{\textcolor{stage1color}{[1]}} Relevant tables: \texttt{country} with columns Code, Name, Continent, Region, Population, \ldots

Skill applied:
\begin{itemize}[nosep, leftmargin=12pt]
\item \textbf{\textcolor{skillcolor}{filter-with-where-and-nulls}}: to filter countries by continent
\item \textbf{\textcolor{skillcolor}{verify-column-semantics}}: to confirm that ``Continent'' is a valid column and contains continent names
\end{itemize}
\vspace{2pt}
First, I will verify that the ``Continent'' column exists and contains values like `Asia' by checking the distinct values.
\action{SELECT DISTINCT Continent FROM country;}

\vspace{2pt}
\textbf{\textcolor{stage1color}{[2]}} \textit{Obs:} \texttt{South America, Antarctica, Africa, North America, Asia, Europe, Oceania.}

The observation confirms ``Asia'' exists in the \texttt{Continent} column as a human-readable name. Now I can construct the count query with confidence.
\action{SELECT COUNT(*) FROM country WHERE Continent = 'Asia';}

\vspace{2pt}
\textbf{\textcolor{stage1color}{[3]}} \textit{Obs:} \texttt{51.} The query returned 51, consistent with the verified schema.
\action{SELECT COUNT(*) FROM country WHERE Continent = 'Asia';}

\vspace{4pt}
\successbadge\ \small The model leverages \texttt{verify-column-semantics} to confirm the column stores human-readable names before filtering, establishing a verify-then-query habit that prevents errors on ambiguous schemas.
\end{casebox}

\vspace{6pt}
\noindent\textcolor{stage1color}{\rule{3pt}{10pt}}\hspace{4pt}\textbf{Fuzzy Matching Failure Spiral}

\vspace{2pt}
\begin{casebox}[stage1color]
\small
\textbf{Database:} driving\_school \hfill \textbf{Q:} \textit{``What city does Janessa Sawayn live in?''}

\vspace{3pt}
\textbf{\textcolor{stage1color}{[1--5]}} Repeated attempts with LIKE:
\action{SELECT a.city FROM Customers c}
\action{JOIN Addresses a ON c.customer\_address\_id = a.address\_id}
\action{WHERE c.first\_name LIKE '\%Janessa\%'}
\action{~~AND c.last\_name LIKE '\%Sawayn\%';}

\textit{Obs:} Empty result each time. The model retries the same query with minor formatting changes.

\vspace{2pt}
\textbf{\textcolor{stage1color}{[6--7]}} Skill applied:
\begin{itemize}[nosep, leftmargin=12pt]
\item \textbf{\textcolor{skillcolor}{fuzzy-text-matching}}: to handle potential spelling variations in the name
\end{itemize}
\vspace{2pt}
Progressively loosened OR conditions:
\action{SELECT a.city FROM Customers c}
\action{JOIN Addresses a ON c.customer\_address\_id = a.address\_id}
\action{WHERE c.first\_name LIKE '\%Janessa\%'}
\action{~~OR c.first\_name LIKE '\%Jana\%'}
\action{~~OR c.first\_name LIKE '\%Jan\%'}
\action{~~AND (c.last\_name LIKE '\%Sawayn\%'}
\action{~~OR c.last\_name LIKE '\%Sawyn\%');}

\vspace{2pt}
\textbf{\textcolor{stage1color}{[8--10]}} ``I applied the \textbf{\textcolor{skillcolor}{fuzzy-text-matching}} skill to refine the name search for `Janessa Sawayn'. The initial queries failed to return results, possibly due to slight variations in spelling. By expanding the search to include similar first names (e.g., `Jana', `Jan') and related last name variations (e.g., `Sawyn')...''

Continues broadening LIKE patterns until the query becomes overly permissive, returning unrelated rows.

\vspace{4pt}
\failbadge\ \small The model over-applies the skill, progressively loosening conditions when the entity simply does not exist in the database. This illustrates a limitation where the base model lacks the capacity to recognize when a skill is inapplicable.
\end{casebox}

\end{document}